\documentclass{article}
\usepackage{graphicx} % Required for inserting images
\usepackage{amsfonts}
\usepackage{caption}
\usepackage{subcaption}
\usepackage{makecell}
\usepackage{tipa}
\usepackage{biblatex}
\title{\textbf{Investigating the Invertibility of Multimodal Latent Spaces: Limitations of Optimization-Based Methods}}
\author{\textbf{Siwoo Park}\\
parkseeuuu@gmail.com}
\date{\today}

\begin{document}
\maketitle
\begin{abstract}
    \noindent This paper investigates the inverse capabilities and broader utility of multimodal latent spaces within task-specific AI (Artificial Intelligence) models. While these models excel at their designed forward tasks (e.g., text-to-image generation, audio-to-text transcription), their potential for inverse mappings remains largely unexplored. We propose an optimization-based framework to infer input characteristics from desired outputs, applying it bidirectionally across Text-Image (BLIP, Flux.1-dev) and Text-Audio (Whisper-Large-V3, Chatterbox-TTS) modalities.
    \\\\
    Our central hypothesis posits that while optimization can guide models towards inverse tasks, their multimodal latent spaces will not consistently support semantically meaningful and perceptually coherent inverse mappings. Experimental results consistently validate this hypothesis. We demonstrate that while optimization can force models to produce outputs that align textually with targets (e.g., a text-to-image model generating an image that an image captioning model describes correctly, or an ASR model transcribing optimized audio accurately), the perceptual quality of these inversions is chaotic and incoherent. Furthermore, when attempting to infer the original semantic input from generative models, the reconstructed latent space embeddings frequently lack semantic interpretability, aligning with nonsensical vocabulary tokens.
    \\\\
    These findings highlight a critical limitation. multimodal latent spaces, primarily optimized for specific forward tasks, do not inherently possess the structure required for robust and interpretable inverse mappings. Our work underscores the need for further research into developing truly semantically rich and invertible multimodal latent spaces.
\end{abstract}
\section{Introduction}
    Rapid advancements in Artificial Intelligence (AI) have significantly enhanced computational capabilities in diverse data domains and modalities. Although task-specific models have shown remarkable performance in their intended forward tasks, their underlying multimodal latent spaces are optimized primarily for these specific functions. Consequently, the full potential of task-specific models, particularly the inverse capabilities and the broader utility of multimodal latent spaces beyond their designed tasks, remains largely unexplored.
\subsection{Research Questions}
    This paper addresses fundamental questions at the intersection of multimodal machine learning and inverse problems.
    \begin{enumerate}
        \item{Can the task-specific models, trained for forward mappings (transforming input data within a specific modality into an output modality), be applied for its inverse tasks (e.g., inferring the characteristics of a text prompt given an image generated by a text-to-image model, or deriving a text prompt that a text-to-audio model might have processed from a generated audio) through optimization-based methods?}
        \item{Can the multimodal latent spaces of task-specific models support a semantically meaningful and perceptually coherent inverse mapping through optimization-based methods?}
    \end{enumerate}
\subsection{Hypothesis}
Our central hypothesis is that the application of optimization-based methods to the task-specific models will reveal specific capabilities and limitations concerning inverse tasks. We hypothesize the following.
\begin{enumerate}
\item We hypothesize that task-specific models can be applied for its inverse tasks through optimization-based methods.
\item We further hypothesize that the multimodal latent spaces of task-specific models will not consistently support semantically meaningful and perceptually coherent inverse mapping through optimization-based methods. This suggests that multimodal latent spaces, primarily optimized for forward tasks, do not readily support robust and interpretable inverse mappings when pushed beyond their intended forward tasks.
\end{enumerate}
\section{Related Work}
    Rapid growth in the field of Artificial Intelligence (AI) has led to sophisticated models capable of excelling in various tasks and modalities. Our work leverages this progress by investigating the invertibility of multimodal latent spaces. The related work section contextualizes our contribution by reviewing relevant prior researches across key areas: (1) Transfer Learning, (2) Gradient Descent Methods and Optimizers, and (3) Optimization-based Inversion, and (4) Adversarial Attacks.
\subsection{Transfer Learning}
    The inherent ability of machine learning models to generalize and perform tasks beyond their original training scope is a cornerstone of modern AI. This phenomenon is widely explored under the term \textbf{transfer learning}, where the knowledge acquired from solving one problem is applied to a different but related problem.
    \\\\
    Early demonstrations of transfer learning emerged from the success of pre-trained models. In Computer Vision (CV), models pre-trained on large-scale datasets like \textbf{ImageNet} showed that learned feature extractors could be effectively transferred and fine-tuned for diverse vision tasks \cite{deng2009imagenet} \cite{NIPS2012_c399862d}. Similarly, in Natural Language Processing (NLP), the development of \textbf{word embeddings} demonstrated that models trained in vast text data could capture semantic relationships that improved performance on various NLP tasks beyond their original training objectives \cite{mikolov2013efficientestimationwordrepresentations} \cite{inproceedings}.
    \\\\
    The advent of Transformer-based architectures significantly pushed the boundaries of transfer learning. \textbf{Large Language Models (LLM)} such as BERT and the GPT series pre-trained on massive text datasets have achieved state-of-the-art performance at the time, performing a wide array of complex tasks (e.g., summarization, question answering, code generation) \cite{devlin2019bertpretrainingdeepbidirectional} \cite{radford2018improving} \cite{radford2019language}. These researches underscore the capacity of multimodal latent spaces to learn broad knowledge and generalizable reasoning skills.
\subsection{Gradient Descent Methods and Optimizers}
    The success of deep learning fundamentally relies on efficient optimization algorithms, predominantly variants of gradient descent.
    \\\\
    The core principle of \textbf{gradient descent} involves iteratively updating the model parameters in the direction opposite to the gradient of a loss function \cite{cauchy1847methode}. For large datasets, \textbf{stochastic gradient descent (SGD)} and its variants with momentum, became crucial, accelerating convergence by using mini-batches \cite{robbins1951stochastic} \cite{polyak1964some}. The introduction of \textbf{Backpropagation} provided an efficient means to compute these gradients for multilayered neural networks \cite{rumelhart1986learning}.
    \\\\
    Further advancements led to adaptive learning rate optimizers, which dynamically adjust the learning rate for each parameter. Notable examples include \textbf{AdaGrad}, \textbf{RMSprop}, and \textbf{Adam (A Method for Stochastic Optimization)} \cite{duchi2011adaptive} \cite{hinton2012neural} \cite{kingma2017adammethodstochasticoptimization}. Adam, in particular, combines the benefits of RMSprop and momentum, computing adaptive learning rates based on both first and second moments of the gradients, making Adam optimizer robust and widely adopted choice for training diverse deep learning models.
\subsection{Optimization-based Inversion}
    The increasing complexity and widespread adoption of Deep Neural Networks (DNNs) have amplified the need for methodologies that enhance their interpretability and allow deeper insights into their internal workings. Network inversion, a critical technique in this pursuit, focuses on reconstructing input data that would produce specific desired output from a trained model.
    \\\\
    Early approaches to network inversion often involved diverse strategies, including the use of backpropagation and evolutionary algorithms to identify multiple inversion points simultaneously through the highly non-convex loss landscape of the neural network \cite{kindermann1990inversion}.
    \\\\
    Recent work introduced a novel method titled \textbf{Landscape Learning for Neural Network Inversion} \cite{liu2022landscapelearningneuralnetwork}. This work addresses the instability inherent in traditional network inversion by learning a loss landscape where gradient descent becomes significantly more efficient and stable.
\subsection{Adversarial Attacks}
    Although deep learning models have achieved remarkable performance, they are often susceptible to \textbf{adversarial attacks}. These attacks involve making small, often imperceptible, perturbations to the input data that cause a model to misclassify or produce an incorrect output. The existence of adversarial examples highlights vulnerabilities in the robustness of AI models and suggests that their latent spaces may not be as smooth or semantically coherent as intuitively assumed.
    \\\\
    Pioneering work first demonstrated the existence of these adversarial examples \cite{szegedy2014intriguingpropertiesneuralnetworks}. Subsequent research developed various methods to generate such examples. The \textbf{Fast Gradient Sign Method (FGSM)} is a simple yet effective technique that perturbs the input in the direction of the sign of the gradient of the loss function with respect to the input \cite{goodfellow2015explainingharnessingadversarialexamples}. More sophisticated iterative methods include \textbf{Projected Gradient Descent (PGD)} \cite{madry2019deeplearningmodelsresistant}, which applies FGSM iteratively and projects the perturbed input back into a valid range, and the \textbf{Carlini \& Wagner (C\&W) attacks}, which are optimization-based attacks designed to find minimal perturbations \cite{carlini2017evaluatingrobustnessneuralnetworks}.

\subsection{Our Contribution}
    The increasing accessibility of powerful AI models presents both unprecedented opportunities and unique challenges, particularly in understanding the inverse capabilities and inherent limitations within the multimodal latent spaces of task-specific models across diverse data domains. Addressing these challenges, this paper makes the following significant contributions.
    \begin{itemize}
        \item{We propose and implement an optimization-based framework for reverse engineering task-specific models, thus applying to its inverse tasks. While sharing methodological similarities with adversarial attack techniques in leveraging optimization to manipulate inputs for a desired output, our approach uniquely applies these principles to the objective of reverse engineering task-specific models across text, image, and audio}
        \item{Through comprehensive experiments using this framework, we investigate the inverse capabilities and the broader utility of multimodal latent spaces of task-specific models. We demonstrate that while optimization-based methods can guide the input towards a target, the resulting inversions often lack perceptual coherence or semantic interpretability in the target modality. This suggests that the multimodal latent spaces, while highly effective for the model's original task, do not readily support a robust and semantically meaningful inverse mapping, even with powerful optimization techniques. Our findings contribute to a deeper understanding of the nature and limitations of multimodal latent spaces in powerful task-specific models, highlighting the critical need for further research into truly semantically rich and invertible multimodal latent spaces.}
    \end{itemize}
\section{Methodology}
    An optimization problem, in its most general form, involves finding the best solution from a set of all possible solutions. Mathematically, an optimization problem is expressed as follows.
    \begin{equation}
        \min_{x \in S}f(x)
    \end{equation}
    Equation (1) represents the objective of minimizing a function $f(x)$ with respect to a variable $x$, where $x$ must belong to a set $S$.
    \\\\
    We denote a non-convex differentiable function $\textbf{f}: \mathbb{R}^d \rightarrow \mathbb{R}^k$ as a generalized task-specific pre-trained machine learning model, where $d \in \mathbb{Z}^+$ and $k \in \mathbb{Z}^+$. Let $\textbf{x} \in \mathbb{R}^d$ and $\textbf{y} \in \mathbb{R}^k$ be generalized input and output of the model, implying $\textbf{y} = \textbf{f}(\textbf{x})$.
    \\\\
    The goal of model (or network) inversion is to find the optimal $\hat{\textbf{x}} \in \mathbb{R}^d$ that best approximates given $\textbf{y}$, implying $\textbf{y} \approx \textbf{f}(\hat{\textbf{x}})$. By letting a differentiable function $\mathcal{L}: \mathbb{R}^{k} \times \mathbb{R}^k \rightarrow \mathbb{R}$ be a generalized loss (or error) function, we can formally state our problem as an optimization problem.
    \begin{equation}
        \hat{\textbf{x}} = \{ \textbf{x} \mid \mathcal{L}(\textbf{f}(\textbf{x}), \textbf{y}) = \min_{\textbf{x'} \in \mathbb{R}^d}\mathcal{L}(\textbf{f}(\textbf{x'}), \textbf{y}) \}
    \end{equation}
    Equation (2) defines the objective of model (or network) inversion.
    \\\\
    The gradient descent approach is a powerful tool for solving multi-variable optimization problems. This fundamental principle is widely applied, and its effectiveness is further demonstrated by advanced optimization algorithms such as Adam \cite{kingma2017adammethodstochasticoptimization}. Recognizing our optimization problem, we denote $J(\textbf{x}) = \mathcal{L}(\textbf{f}(\textbf{x}), \textbf{y})$ as the objective function. In the gradient descent method, the gradient of the objective function $J(\textbf{x})$ is a vector whose components are the partial derivatives with respect to each variable \cite{cauchy1847methode}. By letting $\textbf{x} = (x_1, x_2, ..., x_d)$, gradient vector of $J(\textbf{x})$ is computed as follows.
    \begin{equation}
        \nabla(J(\textbf{x})) = (\frac{\partial J(\textbf{x})}{\partial x_1}, \frac{\partial J(\textbf{x})}{\partial x_2}, ... , \frac{\partial J(\textbf{x})}{\partial x_d})
    \end{equation}
    Equation (3) defines the gradient vector $\nabla(J(\textbf{x}))$ of a multi-variable objective function $J(\textbf{x})$.
    \\\\
    To illustrate the mechanics of the gradient descent algorithm, we present a representative example. The standard gradient descent method iteratively updates the parameter vector $\textbf{x}$ at each timestep $t$. The update rule is defined as follows.
    \begin{equation}
        \textbf{x}^{(t+1)} = \textbf{x}^{(t)} - \eta \nabla J(\textbf{x}^{(t)})
    \end{equation}
    Equation (4) describes the core update rule for the standard gradient descent algorithm.
    \\\\
    where $\eta$ is the learning rate, and $\nabla J(\textbf{x}^{(t)})$ is the gradient vector of the objective function $J(\textbf{x})$, evaluated at the current parameters $\textbf{x}^{(t)} = (x_{1}^{(t)}, x_{2}^{(t)}, ... , x_{d}^{(t)})$. This update can be expressed component-wise for each $x_i$ as follows.
    \begin{equation}
        x_{i}^{(t+1)} = x_{i}^{(t)} - \eta \frac{\partial J(\textbf{x})}{\partial x_i}\Biggm|_{\textbf{x}=\textbf{x}^{(t)}}
    \end{equation}
    Equation (5) provides the component-wise update rule for the gradient descent algorithm.
    \\\\
    By integrating various optimization approaches, where the input $\textbf{x}$ serves as the adjustable parameter, our primary objective is to accurately approximate a meaningful pseudo-inverse for the generalized model function $\textbf{f}$.
\section{Experiments}
    The experiments are structured around two main areas: \textbf{Text-Image} and \textbf{Text-Audio} modeling. In both areas, we conduct a bidirectional exploration of task-specific models, examining classification model obtained through the reverse engineering of generation model, and generation model constructed from classification model.
\subsection{Text-Image}
    Text-Image section delves into bidirectional text-image modeling, leveraging the potential of the following task-specific models.
    \\\\
    \textbf{BLIP}: BLIP (Bootstrapping Language-Image Pre-training) is a large, pre-trained image-to-text model that has significantly advanced the field of image captioning and broader vision-language tasks \cite{li2022blipbootstrappinglanguageimagepretraining}.
    \\\\
    \textbf{FLUX.1-dev} FLUX-1.dev is a text-to-image generative AI model built on a 12 billion parameter rectified flow transformer architecture \cite{blackforestlabs_2024_github}.
\subsubsection{BLIP in Generation task}
    BLIP is an image-to-text model that processes images in a $384 \times 384$ format. For model inversion, we define the objective function as $J(\textbf{x}) = \mathcal{L}(\textbf{f}(\textbf{x}), \textbf{y})$, where $\textbf{f}: \mathbb{R}^{384 \times 384} \rightarrow \mathbb{R}^k$. The optimization-based framework requires initialization of the input as parameters and calculates gradients with respect to these input parameters to iteratively minimize a chosen loss function, thereby guiding the search for the optimal input. We chose the cross-entropy loss function in the BLIP case and computed gradients for each initialized parameters via Pytorch autograd functionality, and finally report optimization results for Gaussian noise $N(0, 1)$ and base image initializations, optimized using the Adam and AdamW optimizer, respectively \cite{kingma2017adammethodstochasticoptimization} \cite{loshchilov2019decoupledweightdecayregularization}.
    \begin{figure}[h]
    \captionsetup[subfigure]{labelformat=empty}
    \centering
    \begin{subfigure}[h]{0.19\textwidth}
    \centering
    \includegraphics[width=\linewidth]{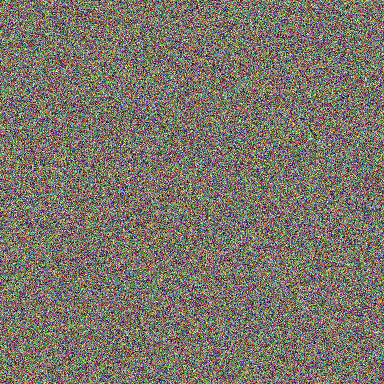}
    \caption{step 0}
    \end{subfigure}
    \begin{subfigure}[h]{0.19\textwidth}
    \centering
    \includegraphics[width=\linewidth]{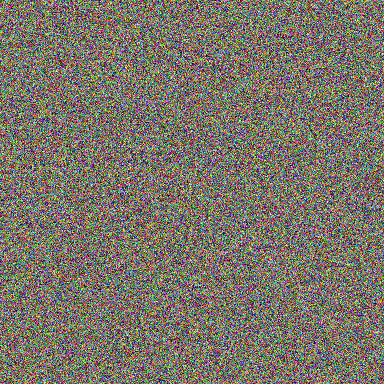}
    \caption{step 10}
    \end{subfigure}
    \begin{subfigure}[h]{0.19\textwidth}
    \centering
    \includegraphics[width=\linewidth]{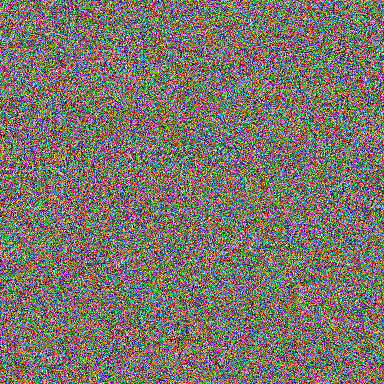}
    \caption{step 100}
    \end{subfigure}
    \begin{subfigure}[h]{0.19\textwidth}
    \centering
    \includegraphics[width=\linewidth]{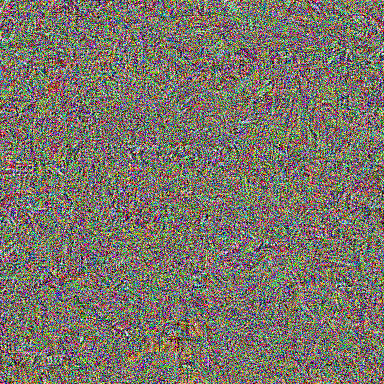}
    \caption{step 1000}
    \end{subfigure}
    \begin{subfigure}[h]{0.19\textwidth}
    \centering
    \includegraphics[width=\linewidth]{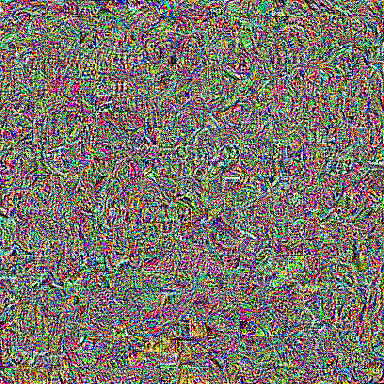}
    \caption{step 10000}
    \end{subfigure}
    \caption{Optimization results using Adam with Gaussian noise initialization for targeting "A red apple on a wooden table."}
    \label{BLIP-Inversion-Adam-Random}
    \end{figure}
    \begin{table}[h]
    \begin{center}
    \caption{Inference for each optimization step}
    \label{Inference-BLIP-Inversion-Adam_Random}
        \begin{tabular}{l|c}
            \textbf{Step} & \textbf{Inference Output}\\
            \hline
            step 0 & this is an image of a television screen with a red background \\
            step 10 & an image of a green background with small squares \\
            step 100 & a red apple on a wooden table \\
            step 1000 & a red apple on a wooden table \\
            step 10000 & a red apple on a wooden table \\
            \end{tabular}
    \end{center}
    \end{table}
    \begin{figure}[h]
    \captionsetup[subfigure]{labelformat=empty}
    \centering
    \begin{subfigure}[h]{0.19\textwidth}
    \centering
    \includegraphics[width=\linewidth]{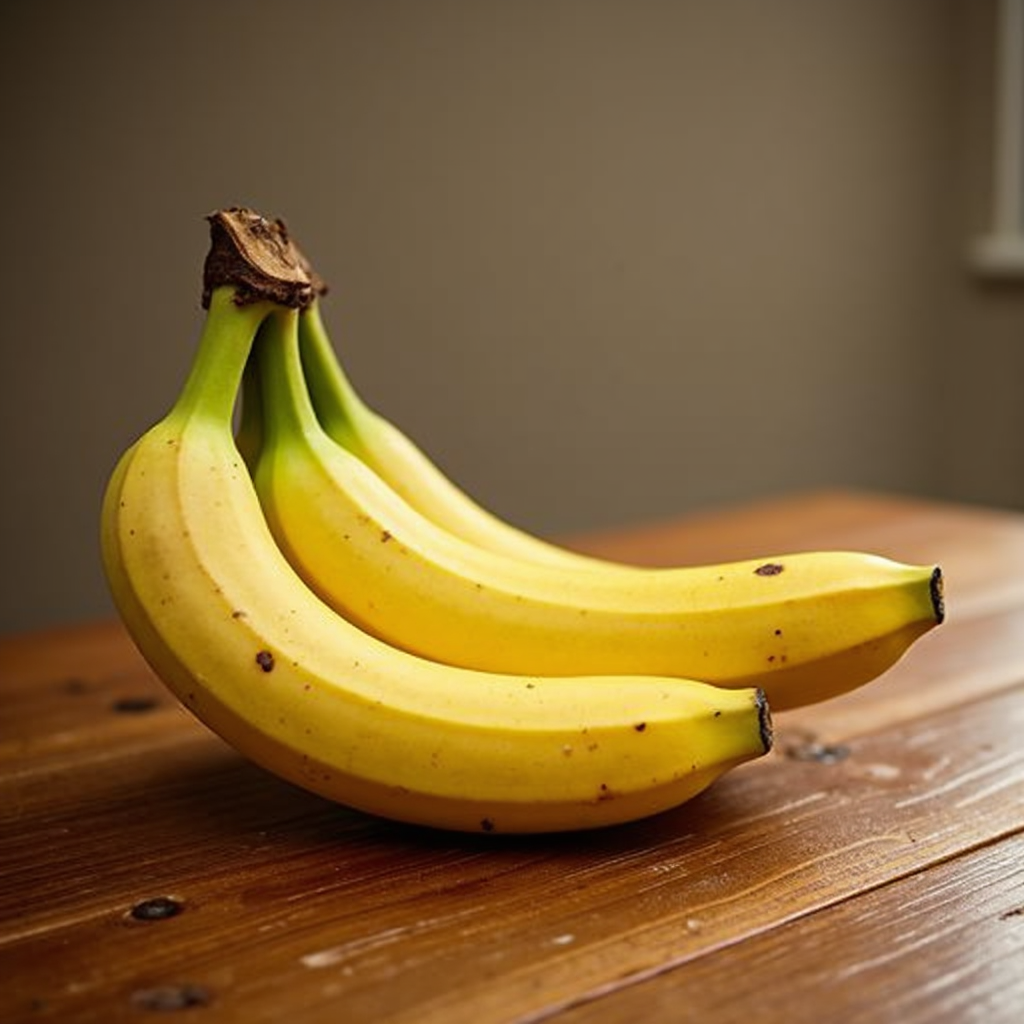}
    \caption{step 0}
    \end{subfigure}
    \begin{subfigure}[h]{0.19\textwidth}
    \centering
    \includegraphics[width=\linewidth]{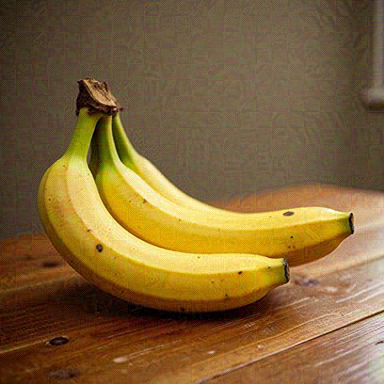}
    \caption{step 10}
    \end{subfigure}
    \begin{subfigure}[h]{0.19\textwidth}
    \centering
    \includegraphics[width=\linewidth]{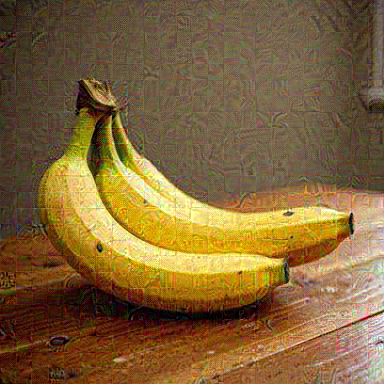}
    \caption{step 100}
    \end{subfigure}
    \begin{subfigure}[h]{0.19\textwidth}
    \centering
    \includegraphics[width=\linewidth]{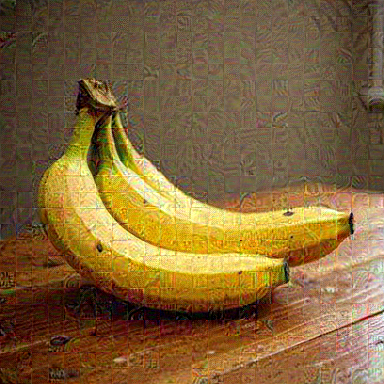}
    \caption{step 1000}
    \end{subfigure}
    \begin{subfigure}[h]{0.19\textwidth}
    \centering
    \includegraphics[width=\linewidth]{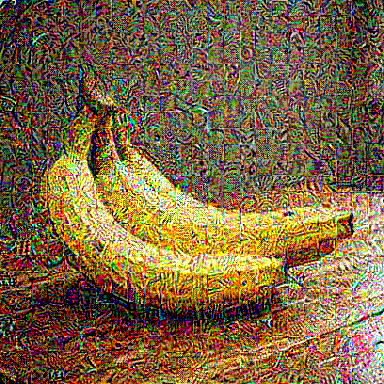}
    \caption{step 10000}
    \end{subfigure}
    \caption{Optimization results using Adam with base image initialization for targeting "A red apple on a wooden table."}
    \label{BLIP-Inversion-Adam-Fix}
    \end{figure}
    \begin{table}[h]
    \begin{center}
    \caption{Inference for each optimization step}
    \label{Inference-BLIP-Inversion-Adam_Fix}
        \begin{tabular}{l|c}
            \textbf{Step} & \textbf{Inference Output}\\
            \hline
            step 0 & there is a bunch of bananas sitting on a wooden table \\
            step 10 & there is a bunch of bananas sitting on a wooden table \\
            step 100 & a red apple on a wooden table \\
            step 1000 & a red apple on a wooden table \\
            step 10000 & a red strawberry on a wooden table \\
            \end{tabular}
    \end{center}
    \end{table}
    \begin{figure}[h]
    \captionsetup[subfigure]{labelformat=empty}
    \centering
    \begin{subfigure}[h]{0.19\textwidth}
    \centering
    \includegraphics[width=\linewidth]{blip_rand_original.png}
    \caption{step 0}
    \end{subfigure}
    \begin{subfigure}[h]{0.19\textwidth}
    \centering
    \includegraphics[width=\linewidth]{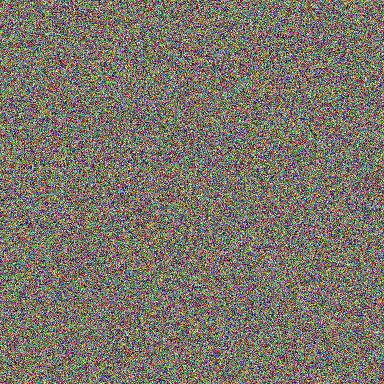}
    \caption{step 10}
    \end{subfigure}
    \begin{subfigure}[h]{0.19\textwidth}
    \centering
    \includegraphics[width=\linewidth]{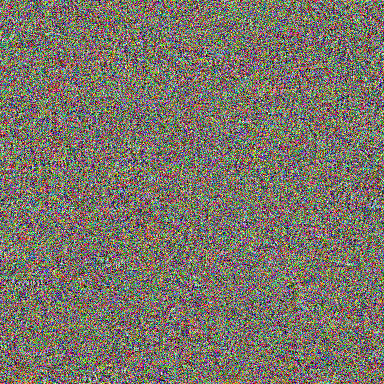}
    \caption{step 100}
    \end{subfigure}
    \begin{subfigure}[h]{0.19\textwidth}
    \centering
    \includegraphics[width=\linewidth]{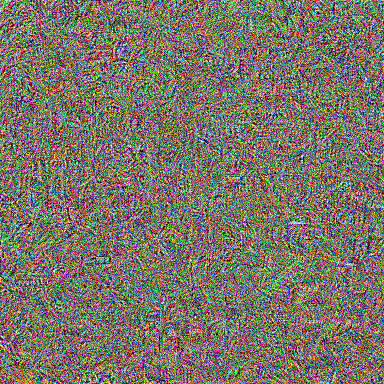}
    \caption{step 1000}
    \end{subfigure}
    \begin{subfigure}[h]{0.19\textwidth}
    \centering
    \includegraphics[width=\linewidth]{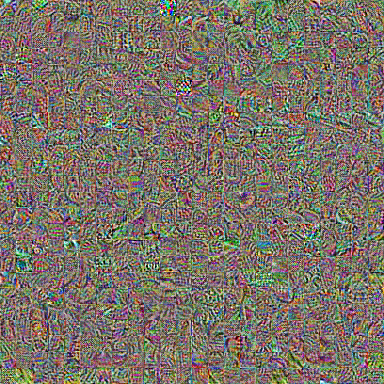}
    \caption{step 10000}
    \end{subfigure}
    \caption{Optimization results using AdamW with Gaussian noise initialization for targeting "A red apple on a wooden table."}
    \label{BLIP-Inversion-AdamW-Random}
    \end{figure}
    \begin{table}[h]
    \begin{center}
    \caption{Inference for each optimization step}
    \label{Inference-BLIP-Inversion-AdamW_Random}
        \begin{tabular}{l|c}
            \textbf{Step} & \textbf{Inference Output}\\
            \hline
            step 0 & this is an image of a television screen with a red background \\
            step 10 & an image of a green background with small dots \\
            step 100 & a red apple on a wooden table \\
            step 1000 & a red apple on a wooden table \\
            step 10000 & a red apple on a wooden table \\
            \end{tabular}
    \end{center}
    \end{table}
    \clearpage
    \begin{figure}[h]
    \captionsetup[subfigure]{labelformat=empty}
    \centering
    \begin{subfigure}[h]{0.19\textwidth}
    \centering
    \includegraphics[width=\linewidth]{banana.png}
    \caption{step 0}
    \end{subfigure}
    \begin{subfigure}[h]{0.19\textwidth}
    \centering
    \includegraphics[width=\linewidth]{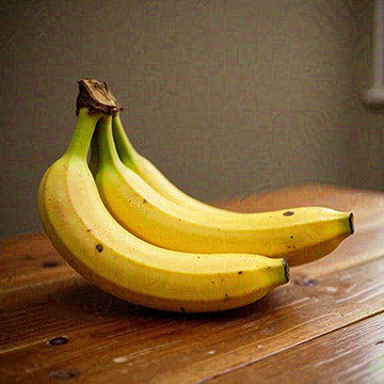}
    \caption{step 10}
    \end{subfigure}
    \begin{subfigure}[h]{0.19\textwidth}
    \centering
    \includegraphics[width=\linewidth]{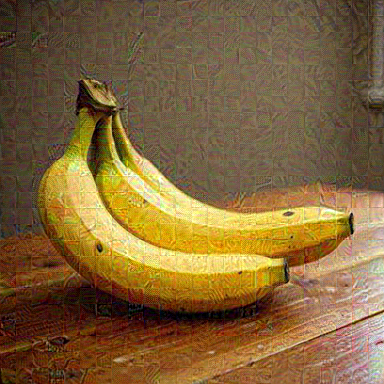}
    \caption{step 100}
    \end{subfigure}
    \begin{subfigure}[h]{0.19\textwidth}
    \centering
    \includegraphics[width=\linewidth]{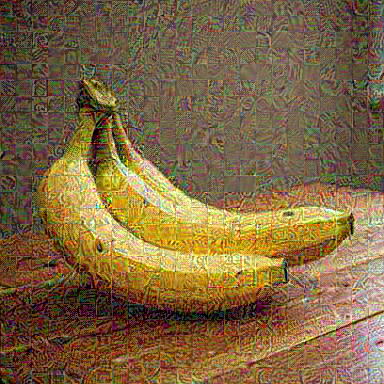}
    \caption{step 1000}
    \end{subfigure}
    \begin{subfigure}[h]{0.19\textwidth}
    \centering
    \includegraphics[width=\linewidth]{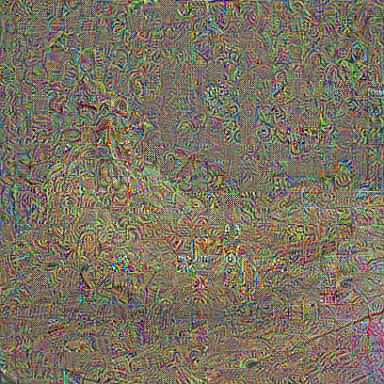}
    \caption{step 10000}
    \end{subfigure}
    \caption{Optimization results using AdamW with base image initialization for targeting "A red apple on a wooden table."}
    \label{BLIP-Inversion-AdamW-Fix}
    \end{figure}
    \begin{table}[h]
    \begin{center}
    \caption{Inference for each optimization step}
    \label{Inference-BLIP-Inversion-AdamW_Fix}
        \begin{tabular}{l|c}
            \textbf{Step} & \textbf{Inference Output}\\
            \hline
            step 0 & there is a bunch of bananas sitting on a wooden table \\
            step 10 & there is a bunch of bananas on a wooden table \\
            step 100 & a red apple on a wooden table \\
            step 1000 & a red apple on a wooden table \\
            step 10000 & a red apple on a wooden table \\
            \end{tabular}
    \end{center}
    \end{table}
    \noindent Each image in Figure 1-4 is processed by BLIP, with the generated output presented in Tables 1-4 respectively.
\subsubsection{Flux.1-dev in Classification task}
    Flux.1-dev model operates as a text-to-image model, mapping textual descriptions to visual output. For computational efficiency and resource optimization, we utilize a 4-bit quantized version of the model. The images are generated at a resolution of $256 \times 256$ pixels. The textual input is processed with a maximum sequence length of 10 tokens. Each token is represented by a 4096-dimensional prompt embedding, while the entire prompt is summarized by a 768-dimensional pooled prompt embedding. Given the use of an empty string for classifier-free guidance, the forward pass of the model can be formally defined as a function $\textbf{f}: \mathbb{R}^{10 \times 4096} \times \mathbb{R}^{768} \rightarrow \mathbb{R}^{256 \times 256}$. While typical diffusion models require multiple iterative denoising steps for image generation, our objective is to efficiently derive the text representation (latent space) from a given image. To achieve our objective, we focus on a single-step inference. Our objective function for this task is formulated as $J(\textbf{x}) = \mathcal{L}(\textbf{f}(\textbf{x}), \textbf{y})$, where $\textbf{x}$ represents the text embeddings (both the token embeddings and the pooled prompt embeddings), $\textbf{y}$ is the target image, and $\mathcal{L}$ denotes a suitable loss function. This objective aims to yield effective approximations for the text embeddings that correspond to the input image. We computed the gradients for initialized input via Pytorch autograd functionality and propose the result of our work on Flux.1-dev.
    \clearpage
    \begin{figure}[h]
    \captionsetup[subfigure]{labelformat=empty}
    \centering
    \begin{subfigure}[h]{0.19\textwidth}
    \centering
    \includegraphics[width=\linewidth]{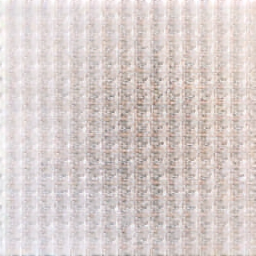}
    \caption{step 0}
    \end{subfigure}
    \begin{subfigure}[h]{0.19\textwidth}
    \centering
    \includegraphics[width=\linewidth]{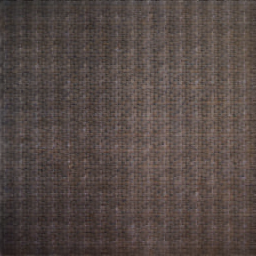}
    \caption{step 25}
    \end{subfigure}
    \begin{subfigure}[h]{0.19\textwidth}
    \centering
    \includegraphics[width=\linewidth]{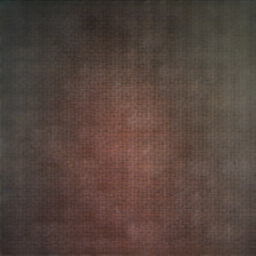}
    \caption{step 50}
    \end{subfigure}
    \begin{subfigure}[h]{0.19\textwidth}
    \centering
    \includegraphics[width=\linewidth]{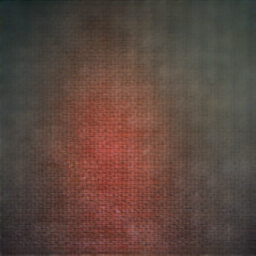}
    \caption{step 75}
    \end{subfigure}
    \begin{subfigure}[h]{0.19\textwidth}
    \centering
    \includegraphics[width=\linewidth]{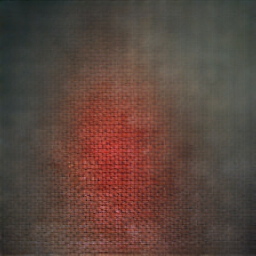}
    \caption{step 100}
    \end{subfigure}
    \begin{subfigure}[h]{0.19\textwidth}
    \centering
    \includegraphics[width=\linewidth]{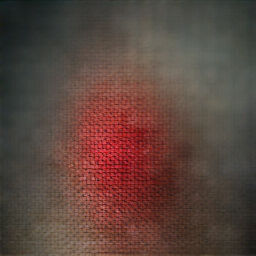}
    \caption{step 125}
    \end{subfigure}
    \begin{subfigure}[h]{0.19\textwidth}
    \centering
    \includegraphics[width=\linewidth]{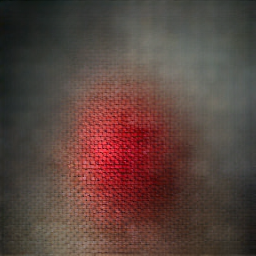}
    \caption{step 150}
    \end{subfigure}
    \begin{subfigure}[h]{0.19\textwidth}
    \centering
    \includegraphics[width=\linewidth]{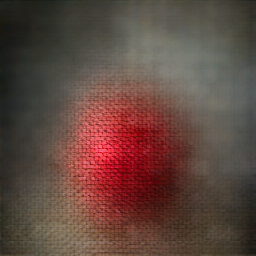}
    \caption{step 175}
    \end{subfigure}
    \begin{subfigure}[h]{0.19\textwidth}
    \centering
    \includegraphics[width=\linewidth]{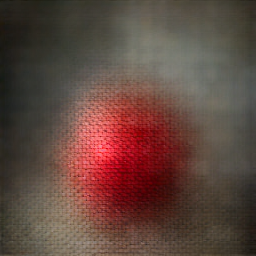}
    \caption{step 200}
    \end{subfigure}
    \begin{subfigure}[h]{0.19\textwidth}
    \centering
    \includegraphics[width=\linewidth]{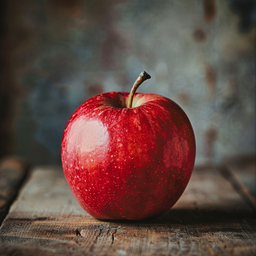}
    \caption{target}
    \end{subfigure}
    \caption{Single-step Inference Optimization using AdamW with Gaussian noise initialization for target image.}
    \label{FLUX-Training-Image}
    \end{figure}
    \noindent We optimized an input represented by a tensor of shape $\mathbb{R}^{10 \times 4096}$ concatenated with a vector of shape $\mathbb{R}^{768}$, using AdamW optimizer to minimize the Mean Squared Error (MSE) loss of single-step inference against a target image \cite{loshchilov2019decoupledweightdecayregularization}. The optimization commenced with a Gaussian noise initialization of the input.
    \\\\
    To evaluate the optimization outcomes, specifically how the model reconstructs text from noisy latent space, we performed inference across a range of training steps with optimized input. Each inference was executed with $50$ denoising steps, employing an empty string for classifer-free guidance. Additionally, a guidance scale of $3.5$ was applied to modulate the influence of the conditioning signal.
    \begin{figure}[h]
    \captionsetup[subfigure]{labelformat=empty}
    \centering
    \begin{subfigure}[h]{0.19\textwidth}
    \centering
    \includegraphics[width=\linewidth]{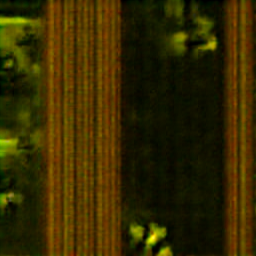}
    \caption{step 0}
    \end{subfigure}
    \begin{subfigure}[h]{0.19\textwidth}
    \centering
    \includegraphics[width=\linewidth]{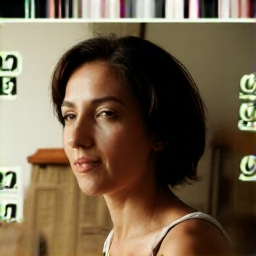}
    \caption{step 25}
    \end{subfigure}
    \begin{subfigure}[h]{0.19\textwidth}
    \centering
    \includegraphics[width=\linewidth]{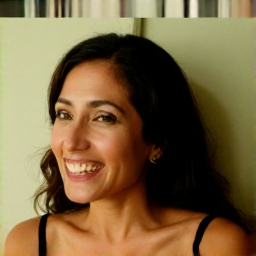}
    \caption{step 50}
    \end{subfigure}
    \begin{subfigure}[h]{0.19\textwidth}
    \centering
    \includegraphics[width=\linewidth]{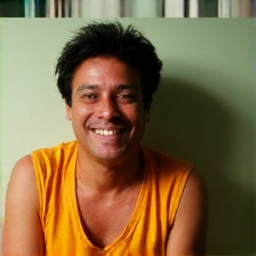}
    \caption{step 75}
    \end{subfigure}
    \begin{subfigure}[h]{0.19\textwidth}
    \centering
    \includegraphics[width=\linewidth]{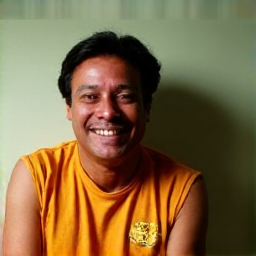}
    \caption{step 100}
    \end{subfigure}
    \begin{subfigure}[h]{0.19\textwidth}
    \centering
    \includegraphics[width=\linewidth]{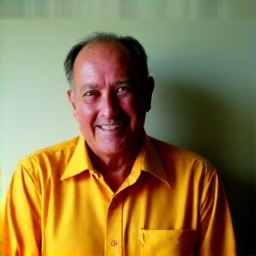}
    \caption{step 125}
    \end{subfigure}
    \begin{subfigure}[h]{0.19\textwidth}
    \centering
    \includegraphics[width=\linewidth]{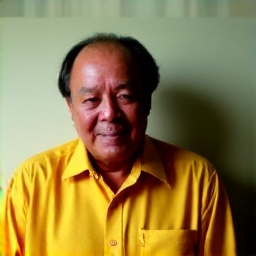}
    \caption{step 150}
    \end{subfigure}
    \begin{subfigure}[h]{0.19\textwidth}
    \centering
    \includegraphics[width=\linewidth]{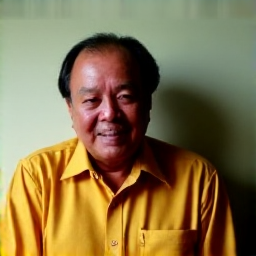}
    \caption{step 175}
    \end{subfigure}
    \begin{subfigure}[h]{0.19\textwidth}
    \centering
    \includegraphics[width=\linewidth]{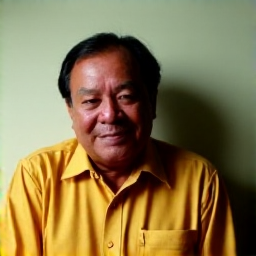}
    \caption{step 200}
    \end{subfigure}
    \begin{subfigure}[h]{0.19\textwidth}
    \centering
    \includegraphics[width=\linewidth]{flux_train_baseline.png}
    \caption{target}
    \end{subfigure}
    \caption{$50$ Denoising Step Inference by Optimization results}
    \label{FLUX-Inference-Image}
    \end{figure}
    \clearpage
    \begin{table}[h]
    \begin{center}
    \caption{Estimated tokens for each step by cosine similarity}
    \label{FLUX-Embedding-Estimation_1}
        \begin{tabular}{|c|c|c|c|c|c|c|c|c|c|c|}
            \hline
            \textbf{Embed} & token 0 & Token 1 & Token 2 & Token 3 & Token 4 \\
            \hline
            step 0 & \makecell{processus \\ 0.0656} & \makecell{purposes \\ 0.0684} & \makecell{Protocol \\ 0.0673} & \makecell{integrate \\ 0.0672} & \makecell{bun \\ 0.0674} \\
            \hline
            step 25 & \makecell{lessness \\ 0.0590} & \makecell{purposes \\ 0.0673} & \makecell{Protocol \\ 0.0688} & \makecell{breach \\ 0.0657} & \makecell{bun \\ 0.0653} \\
            \hline
            step 50 & \makecell{lessness \\ 0.0591} & \makecell{purposes \\ 0.0665} & \makecell{Protocol \\ 0.0683} & \makecell{breach \\ 0.0675} & \makecell{bun \\ 0.0636} \\
            \hline
            step 75 & \makecell{lessness \\ 0.0581} & \makecell{purposes \\ 0.0663} & \makecell{Protocol \\ 0.0684} & \makecell{breach \\ 0.0670} & \makecell{bun \\ 0.0613} \\
            \hline
            step 100 & \makecell{lessness \\ 0.0583} & \makecell{purposes \\ 0.0663} & \makecell{Protocol \\ 0.0684} & \makecell{combinaison \\ 0.0681} & \makecell{unul \\ 0.0621} \\
            \hline
            step 125 & \makecell{lessness \\ 0.0584} & \makecell{purposes \\ 0.0666} & \makecell{Protocol \\ 0.0684} & \makecell{combinaison \\ 0.0687} & \makecell{unul \\ 0.0621} \\
            \hline
            step 150 & \makecell{lessness \\ 0.0589} & \makecell{purposes \\ 0.0668} & \makecell{Protocol \\ 0.0682} & \makecell{combinaison \\ 0.0691} & \makecell{unul \\ 0.0609} \\
            \hline
            step 175 & \makecell{lessness \\ 0.0591} & \makecell{purposes \\ 0.0670} & \makecell{Protocol \\ 0.0682} & \makecell{combinaison \\ 0.0719} & \makecell{unul \\ 0.0590} \\
            \hline
            step 200 & \makecell{lessness \\ 0.0589} & \makecell{purposes \\ 0.0672} & \makecell{Protocol \\ 0.0682} & \makecell{combinaison \\ 0.0730} & \makecell{pamant \\ 0.0585} \\
            \hline
            \end{tabular}
    \end{center}
    \end{table}
    \clearpage
    \begin{table}[h]
    \begin{center}
    \caption{Estimated tokens for each step by cosine similarity}
    \label{FLUX-Embedding-Estimation_2}
        \begin{tabular}
        {|c|c|c|c|c|c|c|c|c|c|c|c|}
            \hline
            \textbf{Embed} & Token 5 & Token 6 & Token 7 & Token 8 & Token 9 & Pooled \\
            \hline
            step 0 & \makecell{Kampf \\ 0.0641} & \makecell{father \\ 0.0781} & \makecell{alter \\ 0.0603} & \makecell{ratio \\ 0.0792} & \makecell{media \\ 0.0588} & \makecell{lina \\ 0.1469} \\
            \hline
            step 25 & \makecell{Kampf \\ 0.0599} & \makecell{father \\ 0.0786} & \makecell{alter \\ 0.0658} & \makecell{ratio \\ 0.0796} & \makecell{media \\ 0.0595} & \makecell{lina \\ 0.1445} \\
            \hline
            step 50 & \makecell{titude \\ 0.0595} & \makecell{father \\ 0.0778} & \makecell{alter \\ 0.0661} & \makecell{ratio \\ 0.0801} & \makecell{media \\ 0.0588} & \makecell{lina \\ 0.1428} \\
            \hline
            step 75 & \makecell{titude \\ 0.0602} & \makecell{father \\ 0.0774} & \makecell{alter \\ 0.0657} & \makecell{ratio \\ 0.0800} & \makecell{media \\ 0.0581} & \makecell{lina \\ 0.1427} \\
            \hline
            step 100 & \makecell{titude \\ 0.0601} & \makecell{father \\ 0.0771} & \makecell{alter \\ 0.0660} & \makecell{ratio \\ 0.0800} & \makecell{RON \\ 0.0595} & \makecell{lina \\ 0.1426} \\
            \hline
            step 125 & \makecell{titude \\ 0.0605} & \makecell{father \\ 0.0769} & \makecell{alter \\ 0.0661} & \makecell{ratio \\ 0.0798} & \makecell{dangerous \\ 0.0599} & \makecell{lina \\ 0.1421} \\
            \hline
            step 150 & \makecell{titude \\ 0.0606} & \makecell{father \\ 0.0770} & \makecell{alter \\ 0.0658} & \makecell{ratio \\ 0.0798} & \makecell{dangerous \\ 0.0605} & \makecell{lina \\ 0.1418} \\
            \hline
            step 175 & \makecell{titude \\ 0.0609} & \makecell{father \\ 0.0774} & \makecell{alter \\ 0.0657} & \makecell{ratio \\ 0.0798} & \makecell{dangerous \\ 0.0605} & \makecell{lina \\ 0.1416} \\
            \hline
            step 200 & \makecell{titude \\ 0.0609} & \makecell{father \\ 0.0777} & \makecell{alter \\ 0.0661} & \makecell{ratio \\ 0.0797} & \makecell{dangerous \\ 0.0597} & \makecell{lina \\ 0.1415} \\
            \hline
            \end{tabular}
    \end{center}
    \end{table}
    \clearpage
    \noindent Each optimized text embedding (input) is processed by Flux.1-dev, with the generated output presented in Figures 5-6.
    \\\\
    We sought to interpret the semantic meaning of our optimized embeddings (input) by estimating their nearest vocabulary tokens. By default, we used the T5 tokenizer for embeddings in the $\mathbb{R}^{10 \times 4096}$ space and the CLIP tokenizer for embeddings in the $\mathbb{R}^{768}$ space. We computed cosine similarity for each embedding against every token within its corresponding tokenizer's vocabulary. The tokens with the highest similarity scores are summarized in Table~\ref{FLUX-Embedding-Estimation_1} and Table~\ref{FLUX-Embedding-Estimation_2}, along with their associated scores, providing insight into the evolving semantics at each inference step. For each single token embedding in $\mathbb{R}^{i}$ form, the cosine similarity score is computed as $\frac{\textbf{A} \cdot \textbf{B}}{||\textbf{A}|| \cdot ||\textbf{B}||}$.
    \begin{figure}[h]
        \centering
        \includegraphics[width=1\textwidth]{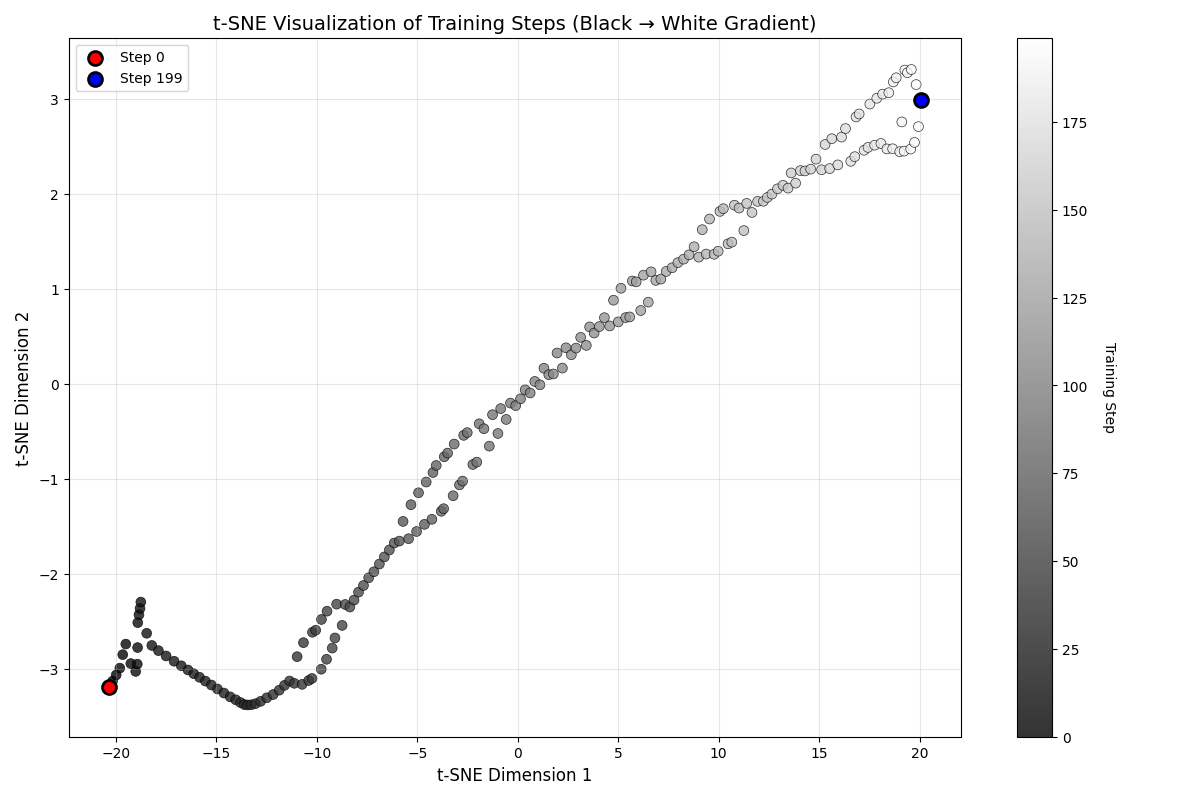}
        \caption{t-SNE $\mathbb{R}^{2}$ visualization of pooled embedding $\mathbb{R}^{768}$ across training steps}
        \label{T-SNE}
    \end{figure}
    \\
    This figure presents a t-SNE projection of the pooled $\mathbb{R}^{768}$ embeddings on $\mathbb{R}^{2}$, which captures their state at different stages of training. The dynamic shifts highlight the model's learning trajectory.
\subsection{Text-Audio}
    Our exploration of bidirectional text-audio modeling is conducted by leveraging the following task-specific models.
    \\\\
    \textbf{Whisper-Large-V3}: Whisper-Large-V3 is OpenAI's advanced automatic speech recognition (ASR) and speech translation model \cite{_2022_whisper} \cite{radford2022robustspeechrecognitionlargescale}. Pre-trained on diverse audio, the model accurately transcribes spoken audio into text across languages and conditions, and translates audio into English. Built on a robust Transformer architecture, the model significantly reduces transcription errors.
    \\\\
    \textbf{Chatterbox-TTS}: Chatterbox-TTS is an open-source, production-grade text-to-speech (TTS) model developed by Resemble AI \cite{resembleai_2025_github}. Using a $0.5$ billion parameter Llama backbone, the model generates highly realistic and expressive speech from text.
\subsubsection{Whisper-Large-V3 in Generation task}
    We utilize the Whisper-Large-V3 model for automatic speech recognition (ASR). The model functions as an audio-to-text mapping, $\textbf{f}: \mathbb{R}^{128 \times 3000} \rightarrow \mathbb{R}^{k}$, which transforms a log-mel spectrogram input into a sequence of text tokens. The input spectrogram is computed from a $30$-second audio clip and consists of 128 Mel frequency bins on $3000$ frames.
    \\\\
    Furthermore, we repurposed the model for text-to-audio (TTA) synthesis. In this investigation, we fix the model's parameters and optimize a randomly initialized (gaussian noise) input audio latent space (the log-mel spectrogram). This optimization aims to minimize the cross-entropy loss with AdamW optimizer between the text transcribed by the model and the target text \cite{loshchilov2019decoupledweightdecayregularization}. The loss between the variable-length generated texts and target texts is computed using an autoregressive objective within the sequence-to-sequence framework of the model. We computed the gradients for initialized input via Pytorch autograd functionality.
    \\\\
    The following figures visualize the log-mel spectrogram across optimization phases.
    \begin{figure}[h]
        \centering
        \includegraphics[width=1\textwidth]{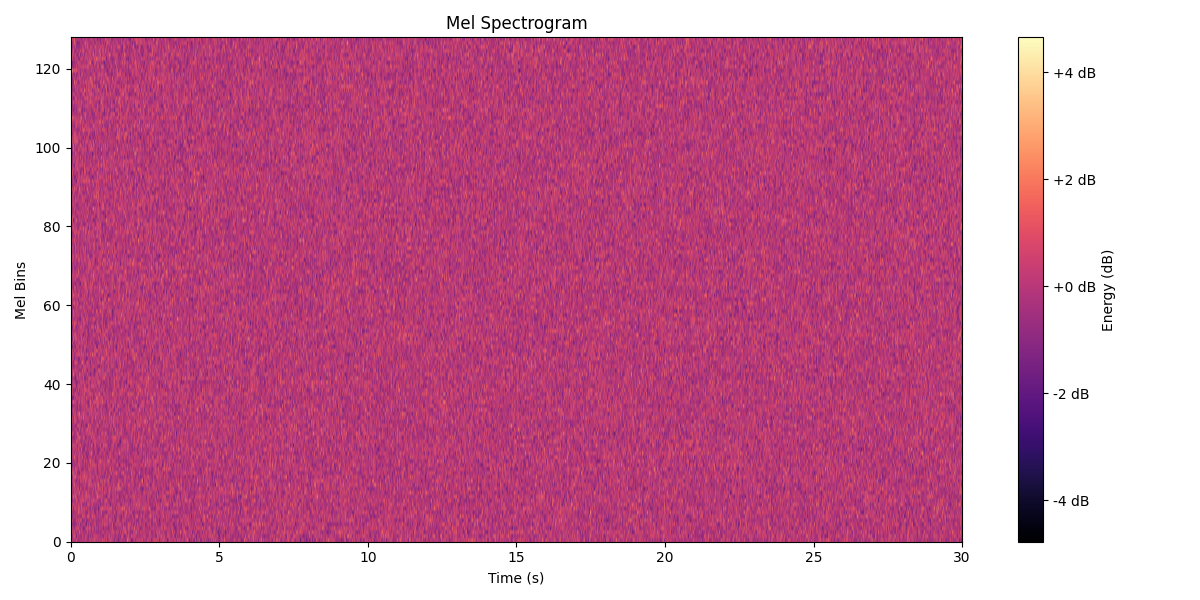}
        \caption{step 0}
        \label{WHISPER-spectrogram_0}
    \end{figure}
    \begin{figure}[h]
        \centering
        \includegraphics[width=1\textwidth]{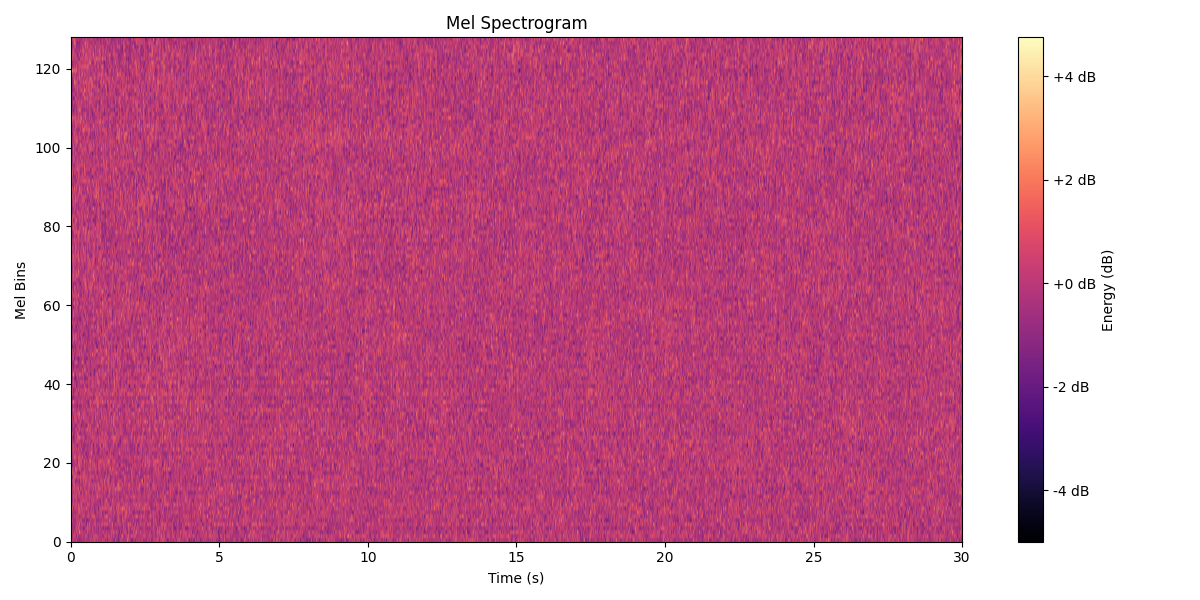}
        \caption{step 750}
        \label{WHISPER-spectrogram_750}
    \end{figure}
    \begin{figure}[h]
        \centering
        \includegraphics[width=1\textwidth]{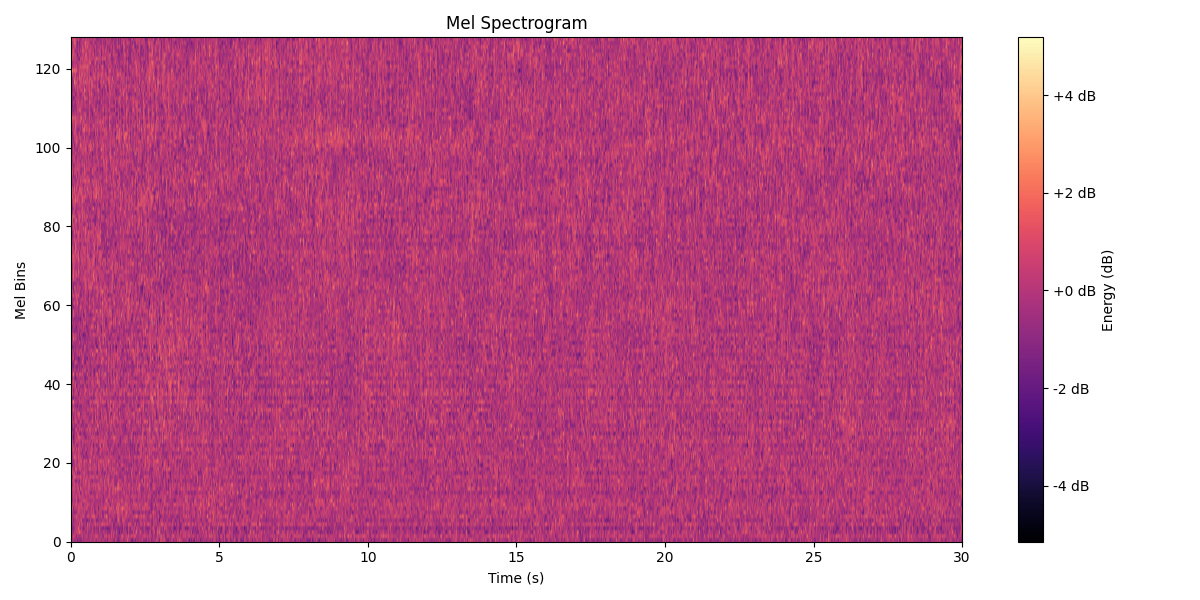}
        \caption{step 1500}
        \label{WHISPER-spectrogram_1500}
    \end{figure}
    \begin{figure}[h]
        \centering
        \includegraphics[width=1\textwidth]{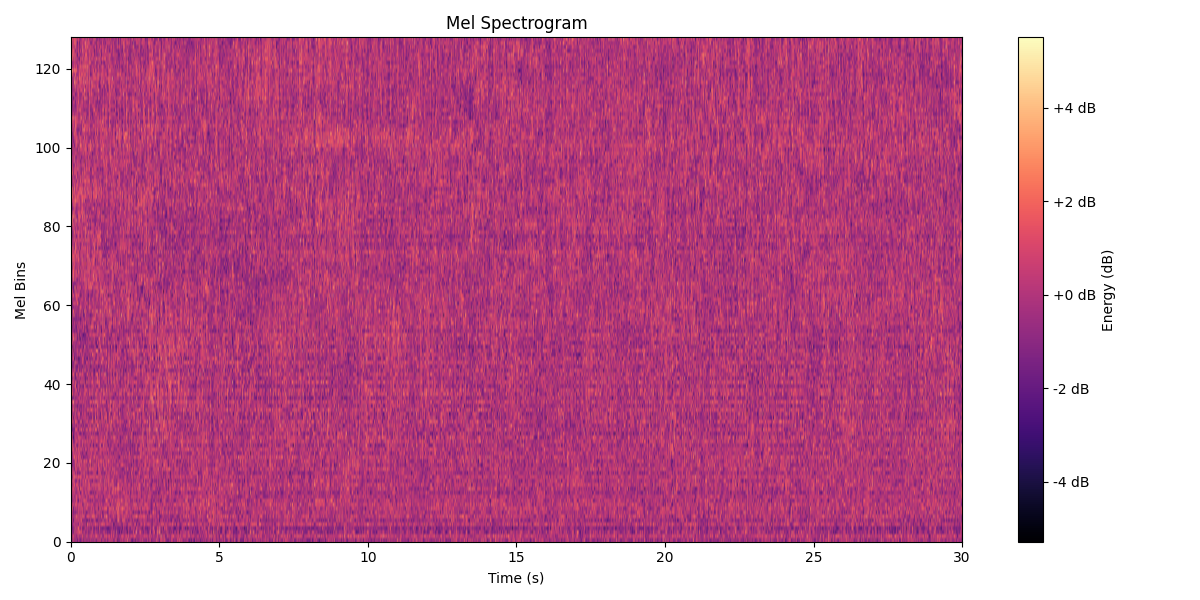}
        \caption{step 2250}
        \label{WHISPER-spectrogram_2250}
    \end{figure}
    \begin{figure}[h]
        \centering
        \includegraphics[width=1\textwidth]{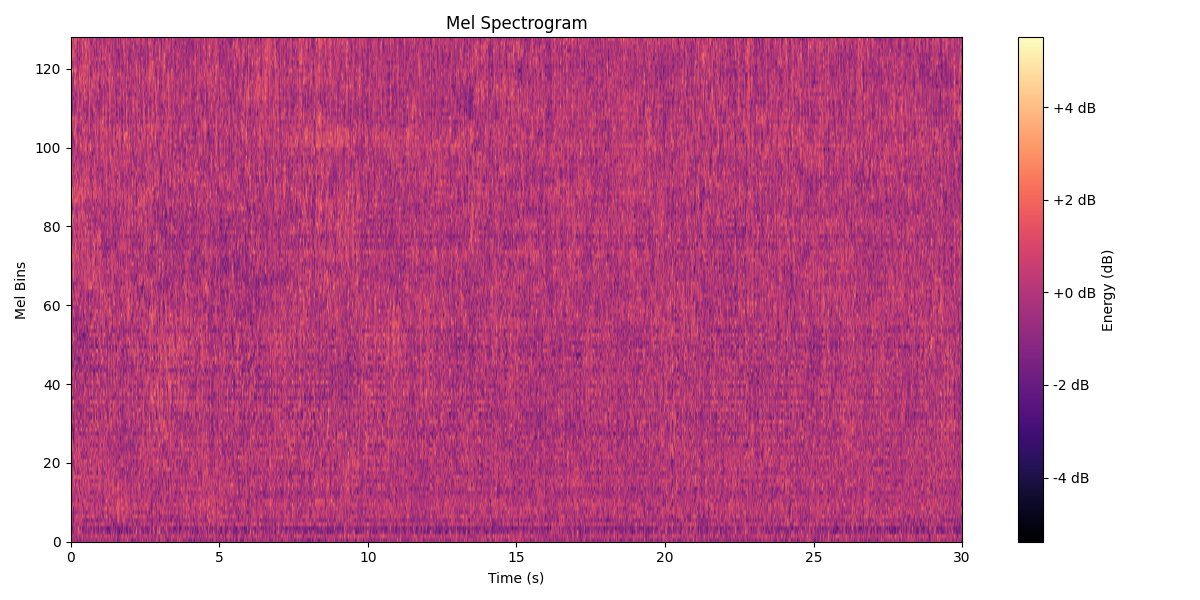}
        \caption{step 3000}
        \label{WHISPER-spectrogram_3000}
    \end{figure}
    \clearpage
    \noindent Figures~\ref{WHISPER-spectrogram_0}-\ref{WHISPER-spectrogram_3000} illustrate the optimization of a $\mathbb{R}^{128 \times 3000}$ audio mel spectrogram for Whisper-Large-V3, aiming to generate the phrase "A red apple on a wooden table". Optimization was performed using the AdamW optimizer, initialized with Gaussian random values \cite{loshchilov2019decoupledweightdecayregularization}.
    \begin{table}[h]
    \begin{center}
    \caption{Inference for each step}
    \label{Inference-WHISPER}
        \begin{tabular}{|c|c|c|}
        \hline
            \textbf{Step} & \textbf{Tokens} & \textbf{Transcription}\\
            \hline
            step 0 & 1 & you\\
            \hline
            step 750 & 113 & \makecell{. . . . . . . . . . . . . . . . . . . . . . . . . . . .\\ . . . . . . . . . . . . . . . . . . . . . . . . . . . .\\ . . . . . . . . . . . . . . . . . . . . . . . . . . . .\\ . . . . . . . . . .. .. .. .. .. .. .. .. .. ..} \\
            \hline
            step 1500 & 62 & \makecell{. . . . . . . . . . . . . . . . . . . . . . .. .. .. .. \\.. .. .. .. red apple on a wooden table. . . .. .\\. .. .. .. .. ...} \\
            \hline
            step 2250 & 22 & . . . . . . . . . . . . . . .. .. .. .. \\
            \hline
            step 3000 & 8 & A red apple on a wooden table.\\
            \hline
            \end{tabular}
    \end{center}
    \end{table}
    \\
    Each optimized spectrogram (input) in Figures 8-12 is processed by Whisper-Large-V3, with the generated output presented in Table 7.
    \\\\
    To demonstrate the effectiveness of the audio log-mel spectrogram optimization, we present the inference results in Table~\ref{Inference-WHISPER}.
    \\\\
    We reconstructed the audio waveform from each optimized log-mel spectrogram using the Griffin-Lim algorithm \cite{griffin1984signal}. The following shows the results of audio reconstruction.
    \begin{figure}[h]
        \centering
        \includegraphics[width=1\textwidth]{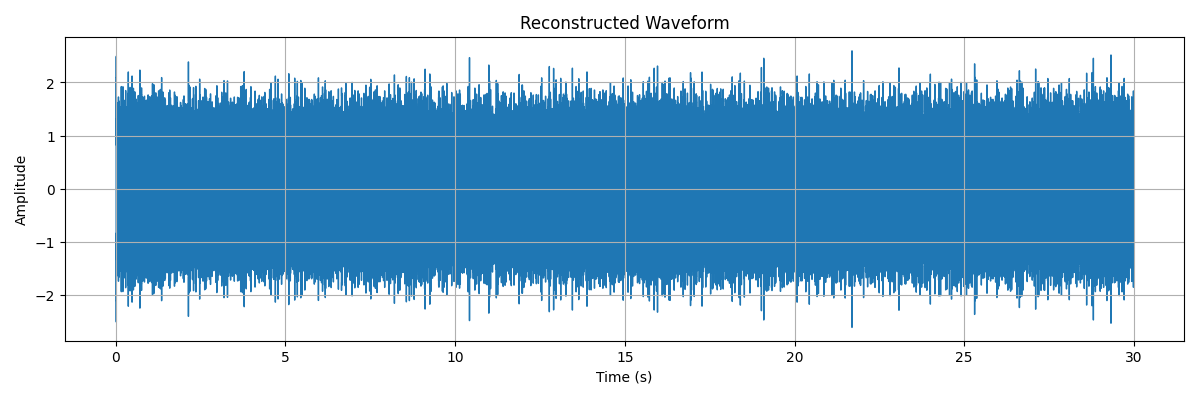}
        \caption{step 0}
        \label{WHISPER-WAVE-0}
    \end{figure}
    \begin{figure}[h]
        \centering
        \includegraphics[width=1\textwidth]{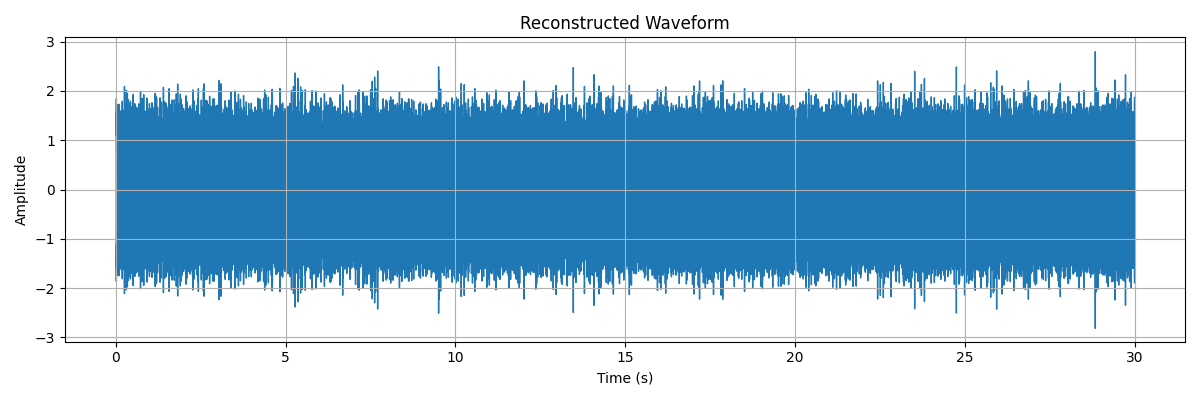}
        \caption{step 750}
        \label{WHISPER-WAVE-750}
    \end{figure}
    \begin{figure}[h]
        \centering
        \includegraphics[width=1\textwidth]{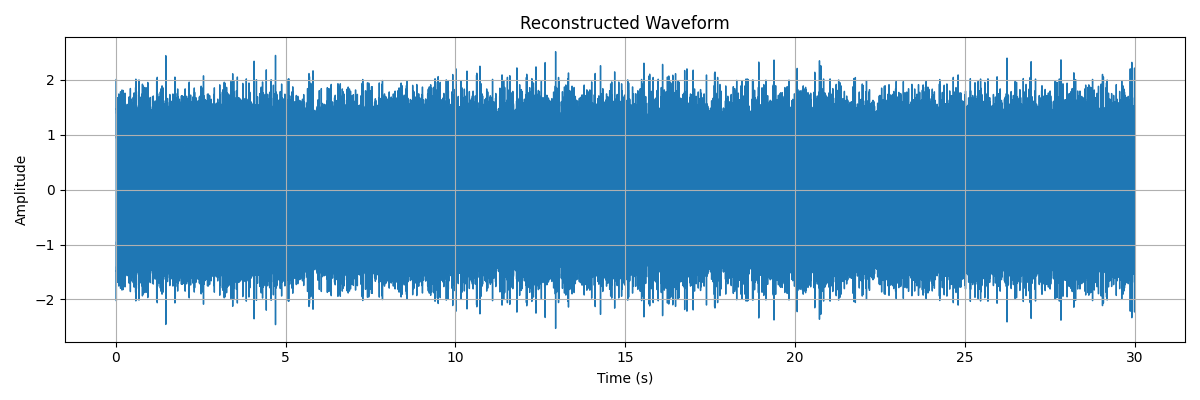}
        \caption{step 1500}
        \label{WHISPER-WAVE-1500}
    \end{figure}
    \begin{figure}[h]
        \centering
        \includegraphics[width=1\textwidth]{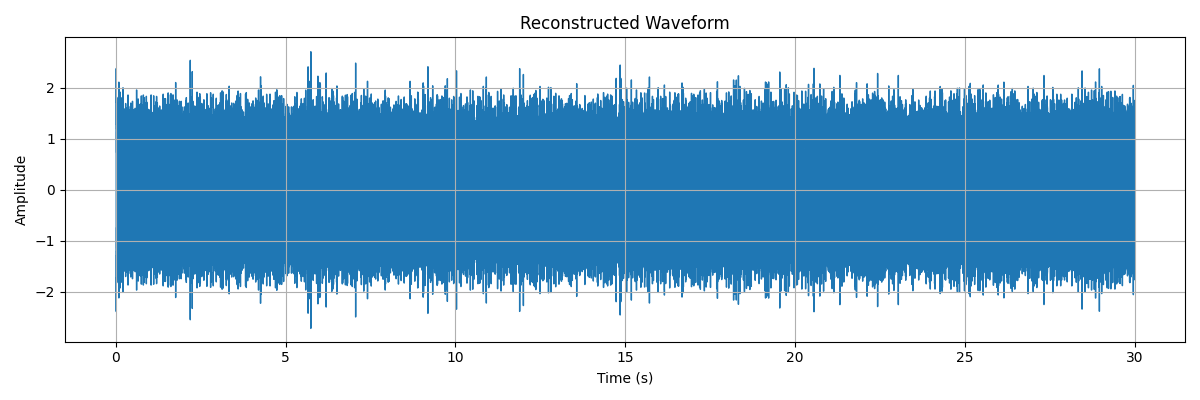}
        \caption{step 2250}
        \label{WHISPER-WAVE-2250}
    \end{figure}
    \begin{figure}[h]
        \centering
        \includegraphics[width=1\textwidth]{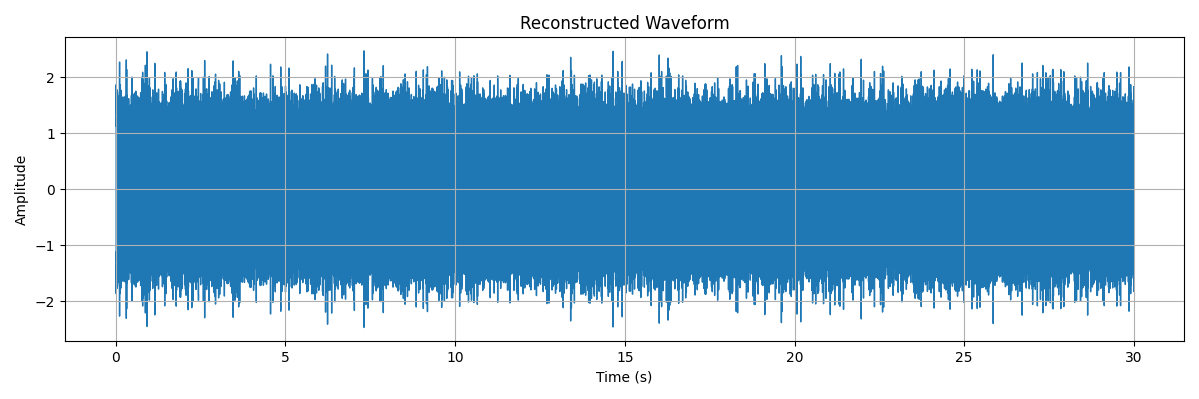}
        \caption{step 3000}
        \label{WHISPER-WAVE-3000}
    \end{figure}
    \clearpage
\subsubsection{Chatterbox-TTS in Classification task}
    The Chatterbox-TTS model synthesizes audio from a sequence of input tokens. Specifically, the model accepts a sequence of $n$ tokens, each represented by a $1024$ dimensional embedding, and generates audio at a $24000$ sample rate. Our experimental objective is to optimize the initial $\mathbb{R}^{n \times 1024}$ dimensional text latent space to precisely generate the desired audio output.
    \\\\
    This work is crucial for understanding the model's sensitivity to input variations and its capacity to produce specific acoustic properties. In our experiments, we fixed the number of tokens $n$ at $23$, used gaussian noise initialization, and optimized for a $53248$ dimensional audio output, which perceptually corresponds to "A red apple on a wooden table". The optimization process heavily relies on the AdamW optimizer and Mel spectrogram loss, which is widely recognized for its effectiveness in evaluating the perceptual similarity of audio signals, particularly in text-to-speech (TTS) and voice-synthesis tasks \cite{loshchilov2019decoupledweightdecayregularization}. We computed the gradients for the initialized input via Pytorch autograd functionality.
    \\\\
    To illustrate the optimization trajectory, we provide visualizations of the mel spectrograms generated throughout the training process.
    \begin{figure}[h]
        \centering
        \includegraphics[width=1\textwidth]{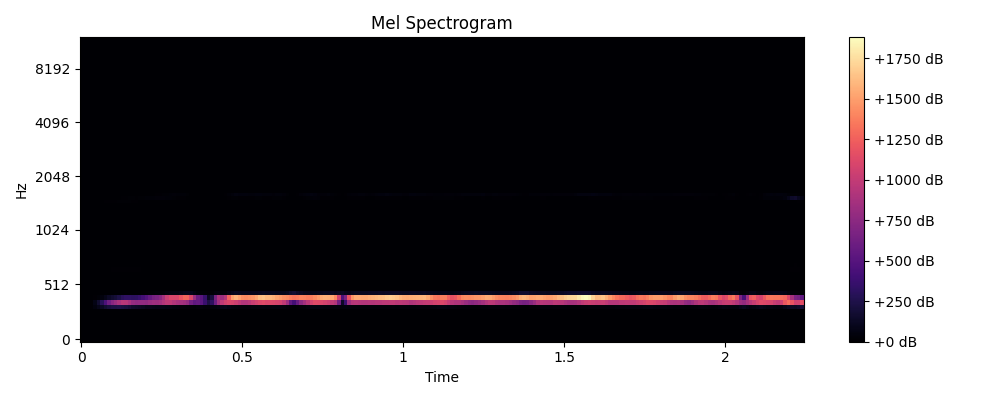}
        \caption{step 0}
        \label{CHAT-SPEC-0}
    \end{figure}
    \begin{figure}[h]
        \centering
        \includegraphics[width=1\textwidth]{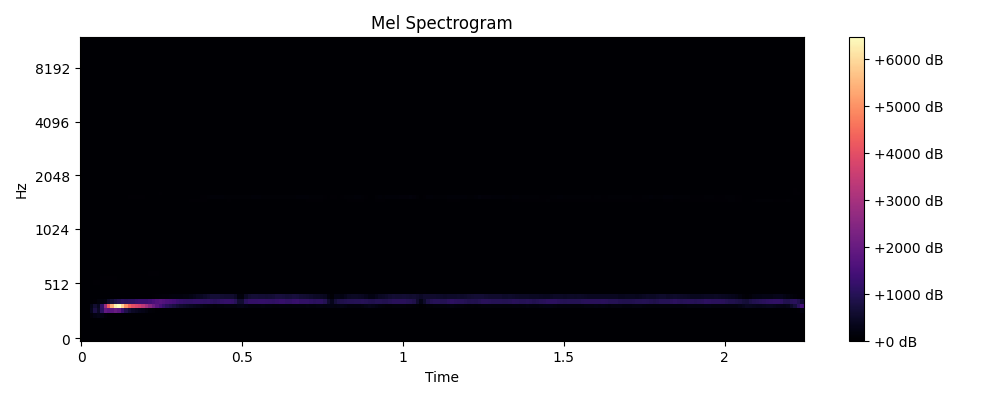}
        \caption{step 250}
        \label{CHAT-SPEC-250}
    \end{figure}
    \begin{figure}[h]
        \centering
        \includegraphics[width=1\textwidth]{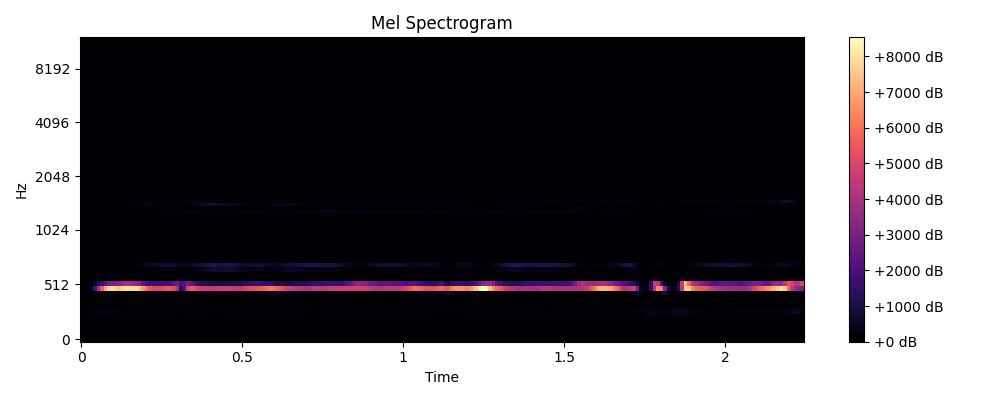}
        \caption{step 500}
        \label{CHAT-SPEC-500}
    \end{figure}
    \begin{figure}[h]
        \centering
        \includegraphics[width=1\textwidth]{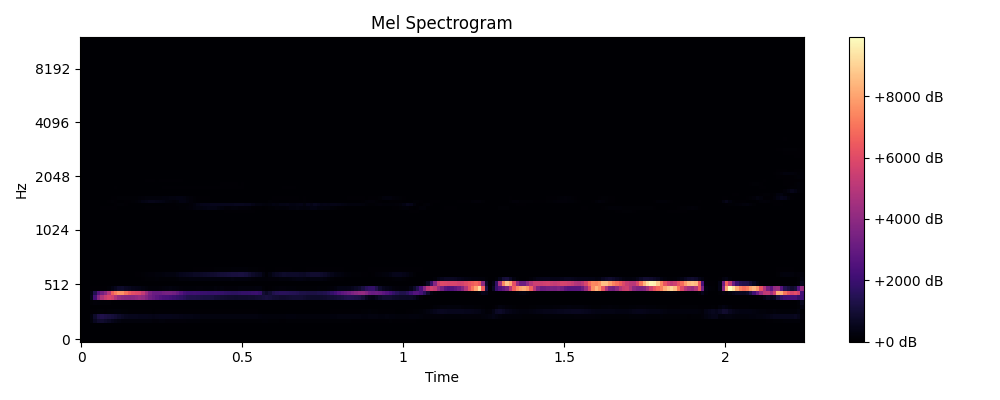}
        \caption{step 750}
        \label{CHAT-SPEC-750}
    \end{figure}
    \begin{figure}[h]
        \centering
        \includegraphics[width=1\textwidth]{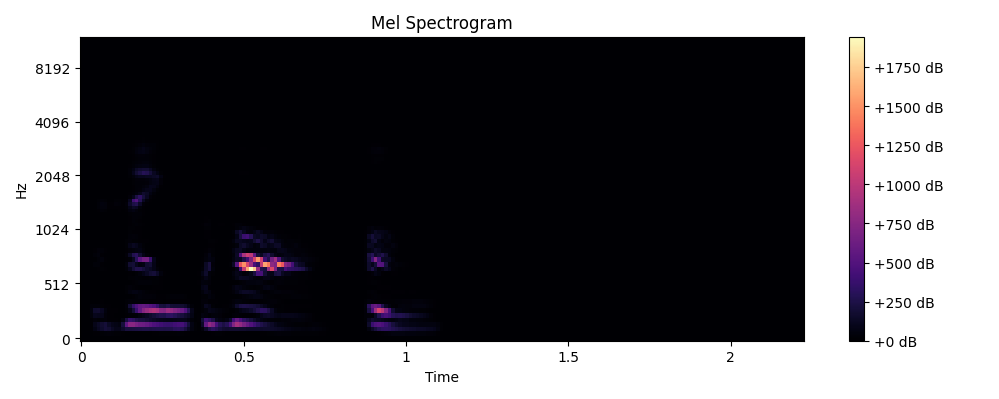}
        \caption{step 1000}
        \label{CHAT-SPEC-1000}
    \end{figure}
    \begin{figure}[h]
        \centering
        \includegraphics[width=1\textwidth]{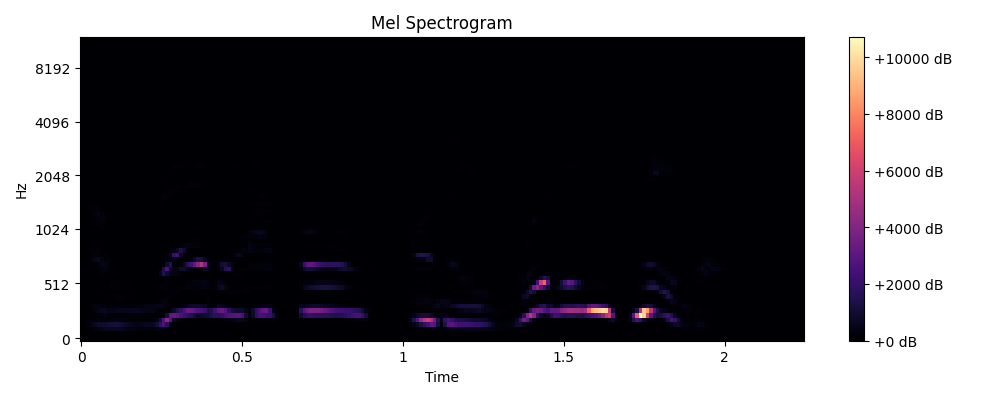}
        \caption{target}
        \label{CHAT-SPEC-BASELINE}
    \end{figure}
    \clearpage
    \noindent To further elucidate the optimization trajectory, we also propose visualizing the synthesized audio waveforms in selected optimization steps.
    \\\\
    Such granular analysis will allow for a direct examination of how the model's output acoustics evolve, complementing the frequency-domain insights provided by the mel spectrograms.
    \begin{figure}[h]
        \centering
        \includegraphics[width=1\textwidth]{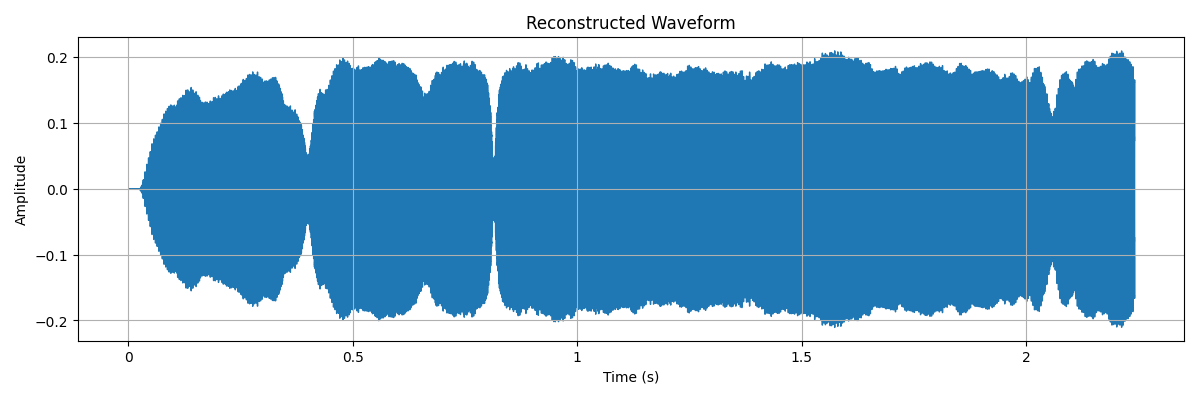}
        \caption{step 0}
        \label{CHAT-WAVEFORM-0}
    \end{figure}
    \begin{figure}[h]
        \centering
        \includegraphics[width=1\textwidth]{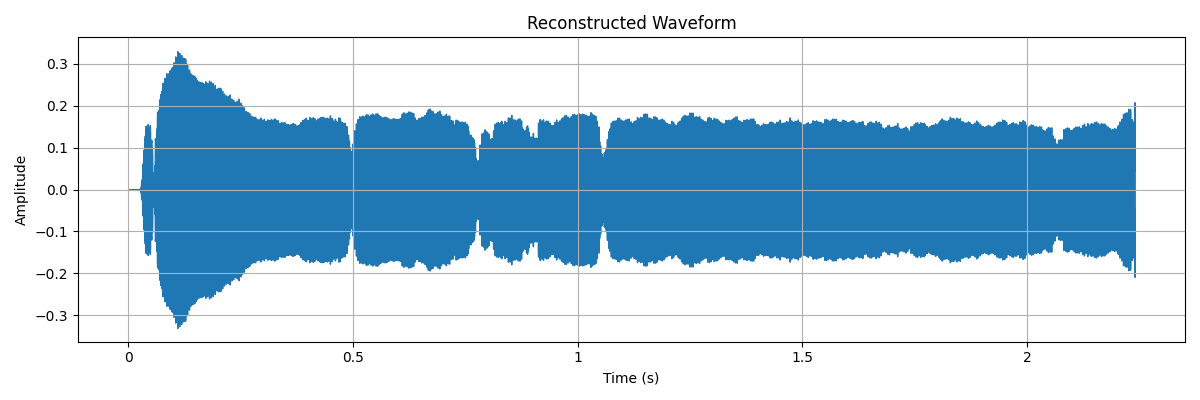}
        \caption{step 250}
        \label{CHAT-WAVEFORM-250}
    \end{figure}
    \begin{figure}[h]
        \centering
        \includegraphics[width=1\textwidth]{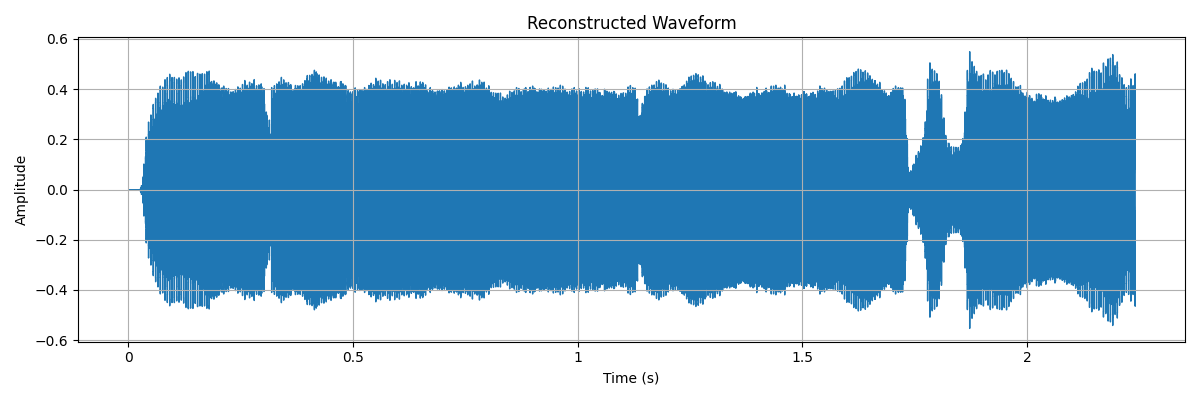}
        \caption{step 500}
        \label{CHAT-WAVEFORM-500}
    \end{figure}
    \begin{figure}[h]
        \centering
        \includegraphics[width=1\textwidth]{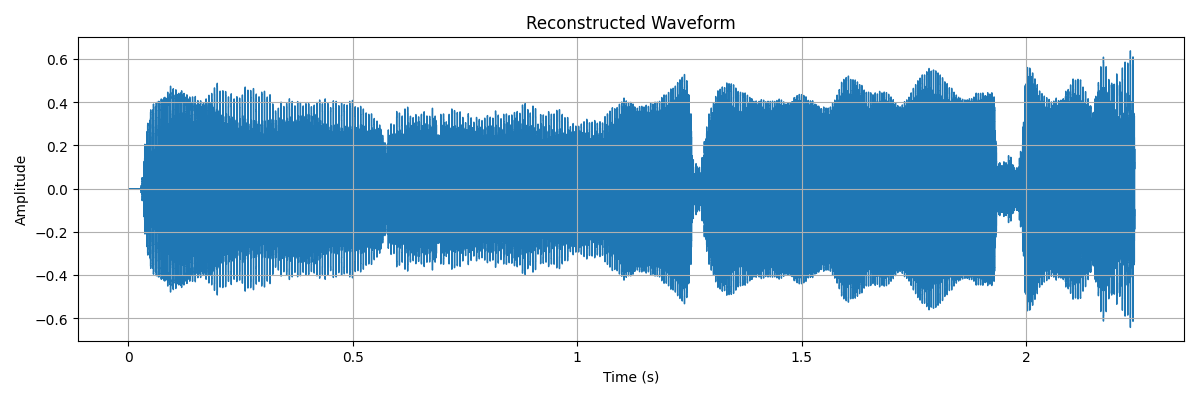}
        \caption{step 750}
        \label{CHAT-WAVEFORM-750}
    \end{figure}
    \begin{figure}[h]
        \centering
        \includegraphics[width=1\textwidth]{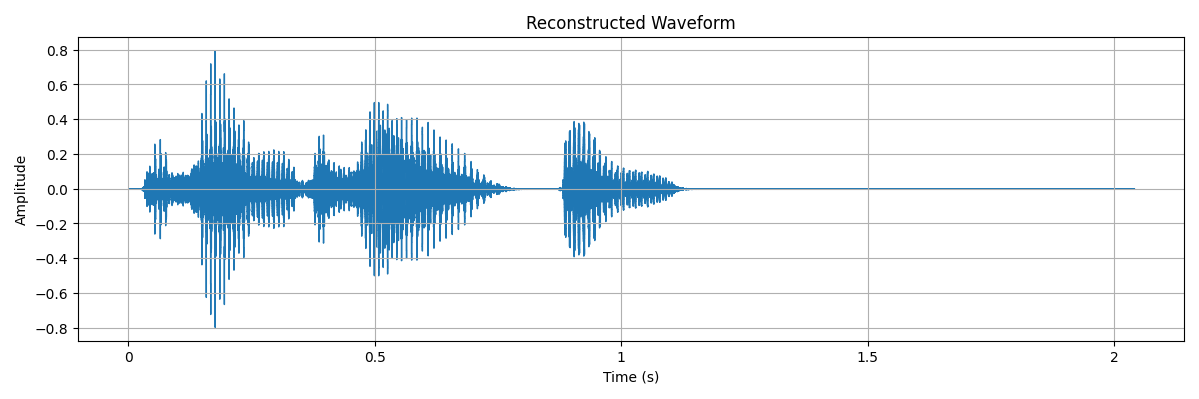}
        \caption{step 1000}
        \label{CHAT-WAVEFORM-1000}
    \end{figure}
    \begin{figure}[h]
        \centering
        \includegraphics[width=1\textwidth]{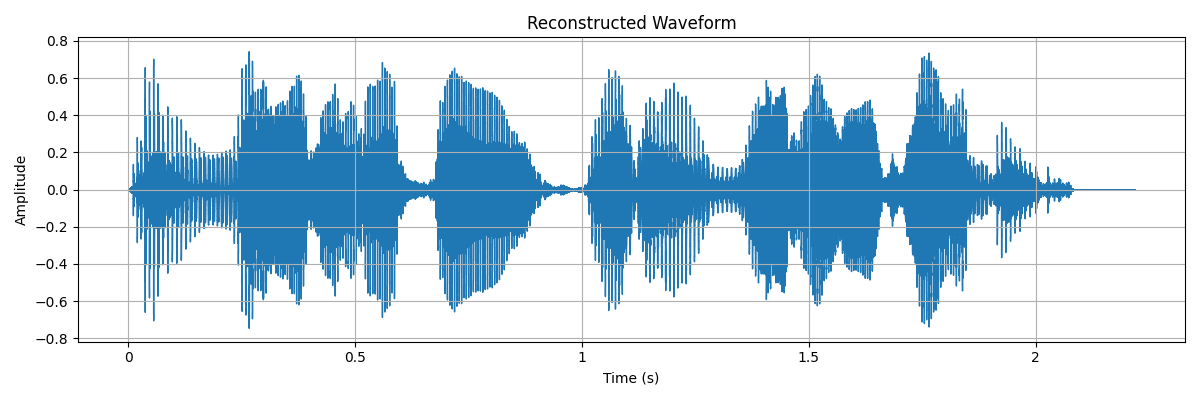}
        \caption{target}
        \label{CHAT-WAVEFORM-BASELINE}
    \end{figure}
    \clearpage
    \noindent After optimizing the embeddings, we performed a cosine similarity analysis to determine the most semantically similar vocabulary token for each optimized embedding. The cosine similarity analysis allowed us to identify which token each optimized embedding implicitly represents within the model's vocabulary. This functions as an interpretative measure of the latent space of the model.
    
    \begin{table}[h]
    \begin{center}
    \caption{Estimated tokens for each step by cosine similarity}
    \label{CHAT-Embedding-Estimation_1}
        \begin{tabular}{|c|c|c|c|c|c|}
            \hline
            \textbf{Embed} & token 0 & Token 1 & Token 2 & Token 3 & Token 4 \\
            \hline
            step 0 & \makecell{˘ \\ 0.1091} & \makecell{\textasciicircum \\ 0.0942} & \makecell{3 \\ 0.0968} & \makecell{\% \\ 0.1023} & \makecell{fr \\ 0.1074} \\
            \hline
            step 250 & \makecell{˘ \\ 0.1091} & \makecell{\textasciicircum \\ 0.0942} & \makecell{3 \\ 0.0968} & \makecell{\% \\ 0.1023} & \makecell{ fr \\ 0.1074} \\
            \hline
            step 500 & \makecell{u\textlengthmark \\ 0.1294} & \makecell{\textsuperscript{y} \\ 0.0856} & \makecell{‐ \\ 0.1326} & \makecell{ca \\ 0.0940} & \makecell{wh \\ 0.1140} \\
            \hline
            step 750 & \makecell{u\textlengthmark \\ 0.1294} & \makecell{\textsuperscript{y} \\ 0.0856} & \makecell{‐ \\ 0.1326} & \makecell{ca \\ 0.0940} & \makecell{wh \\ 0.1140} \\
            \hline
            step 750 & \makecell{u\textlengthmark \\ 0.1294} & \makecell{\textsuperscript{y} \\ 0.0856} & \makecell{‐ \\ 0.1326} & \makecell{ca \\ 0.0940} & \makecell{wh \\ 0.1140} \\
            \hline
            \end{tabular}
    \end{center}
    \end{table}
    \begin{table}[h]
    \begin{center}
    \caption{Estimated tokens for each step by cosine similarity}
    \label{CHAT-Embedding-Estimation_2}
        \begin{tabular}
        {|c|c|c|c|c|c|c|c|c|c|c|c|}
            \hline
            \textbf{Embed} & Token 5 & Token 6 & Token 7 & Token 8 & Token 9 & Token 10 \\
            \hline
            step 0 & \makecell{\textbeltl \\ 0.0978} & \makecell{su \\ 0.0967} & \makecell{ö \\ 0.0829} & \makecell{v \\ 0.0984} & \makecell{Y \\ 0.0922} & \makecell{all \\ 0.0926} \\
            \hline
            step 250 & \makecell{\textbeltl \\ 0.0978} & \makecell{su \\ 0.0967} & \makecell{ö \\ 0.0829} & \makecell{v \\ 0.0984} & \makecell{Y \\ 0.0922} & \makecell{all \\ 0.0926} \\
            \hline
            step 500 & \makecell{ter \\ 0.1021} & \makecell{su \\ 0.0930} & \makecell{who \\ 0.1027} & \makecell{ve \\ 0.0825} & \makecell{? \\ 0.0945} & \makecell{ven \\ 0.0790} \\
            \hline
            step 750 & \makecell{ter \\ 0.1021} & \makecell{su \\ 0.0930} & \makecell{who \\ 0.1027} & \makecell{ve \\ 0.0825} & \makecell{? \\ 0.0945} & \makecell{ven \\ 0.0790} \\
            \hline
            step 750 & \makecell{ter \\ 0.1021} & \makecell{su \\ 0.0930} & \makecell{who \\ 0.1027} & \makecell{ve \\ 0.0825} & \makecell{? \\ 0.0945} & \makecell{ven \\ 0.0790} \\
            \hline
            \end{tabular}
    \end{center}
    \end{table}
    \clearpage
    \begin{table}[h]
    \begin{center}
    \caption{Estimated tokens for each step by cosine similarity}
    \label{CHAT-Embedding-Estimation_3}
        \begin{tabular}
        {|c|c|c|c|c|c|c|c|c|c|c|c|}
            \hline
            \textbf{Embed} & Token 11 & Token 12 & Token 13 & Token 14 & Token 15 & Token 16 \\
            \hline
            step 0 & \makecell{ir \\ 0.0976} & \makecell{§ \\ 0.0933} & \makecell{that \\ 0.0842} & \makecell{\textsuperscript{j} \\ 0.0968} & \makecell{' \\ 0.0859} & \makecell{[sniff] \\ 0.1036} \\
            \hline
            step 250 & \makecell{ir \\ 0.0976} & \makecell{§ \\ 0.0933} & \makecell{that \\ 0.0842} & \makecell{\textsuperscript{j} \\ 0.0968} & \makecell{' \\ 0.0859} & \makecell{[sniff] \\ 0.1036} \\
            \hline
            step 500 & \makecell{\textbari \\ 0.0970} & \makecell{who \\ 0.0941} & \makecell{ent \\ 0.0897} & \makecell{\textsci \\ 0.0839} & \makecell{õ \\ 0.0901} & \makecell{\textsuperscript{y} \\ 0.1198} \\
            \hline
            step 750 & \makecell{\textbari \\ 0.0970} & \makecell{who \\ 0.0941} & \makecell{ent \\ 0.0897} & \makecell{\textsci \\ 0.0839} & \makecell{õ \\ 0.0901} & \makecell{\textsuperscript{y} \\ 0.1198} \\
            \hline
            step 750 & \makecell{\textbari \\ 0.0970} & \makecell{who \\ 0.0941} & \makecell{ent \\ 0.0897} & \makecell{\textsci \\ 0.0839} & \makecell{õ \\ 0.0901} & \makecell{\textsuperscript{y} \\ 0.1198} \\
            \hline
            \end{tabular}
    \end{center}
    \end{table}
    \begin{table}[h]
    \begin{center}
    \caption{Estimated tokens for each step by cosine similarity}
    \label{CHAT-Embedding-Estimation_4}
        \begin{tabular}
        {|c|c|c|c|c|c|c|c|c|c|c|c|}
            \hline
            \textbf{Embed} & Token 17 & Token 18 & Token 19 & Token 20 & Token 21 & Token 22 \\
            \hline
            step 0 & \makecell{ack \\ 0.0952} & \makecell{\textcrh \\ 0.0940} & \makecell{¨ \\ 0.1038} & \makecell{re \\ 0.0978} & \makecell{| \\ 0.0839} & \makecell{al \\ 0.1211} \\
            \hline
            step 250 & \makecell{ack \\ 0.0952} & \makecell{\textcrh \\ 0.0940} & \makecell{¨ \\ 0.1038} & \makecell{re \\ 0.0978} & \makecell{| \\ 0.0839} & \makecell{al \\ 0.1211} \\
            \hline
            step 500 & \makecell{\textminus \\ 0.0980} & \makecell{\textcrh \\ 0.0971} & \makecell{op \\ 0.0860} & \makecell{f \\ 0.0864} & \makecell{[meow] \\ 0.0878} & \makecell{z \\ 0.0867} \\
            \hline
            step 750 & \makecell{\textminus \\ 0.0980} & \makecell{\textcrh \\ 0.0971} & \makecell{op \\ 0.0860} & \makecell{f \\ 0.0864} & \makecell{[meow] \\ 0.0878} & \makecell{z \\ 0.0867} \\
            \hline
            step 1000 & \makecell{\textminus \\ 0.0980} & \makecell{\textcrh \\ 0.0971} & \makecell{op \\ 0.0860} & \makecell{f \\ 0.0864} & \makecell{[meow] \\ 0.0878} & \makecell{z \\ 0.0867} \\
            \hline
            \end{tabular}
    \end{center}
    \end{table}
    \noindent Each optimized text embedding (input) in Tables 8-11 is processed by Chatterbox-TTS, with the generated output presented in Figures 18-22, and Figures 24-28.
\section{Quantitative Consistency Analysis}
    This section presents a detailed quantitative evaluation of the consistency of our results. Our research focused on four distinct task-specific models, each designed for unique applications. For each of these models, we categorized their respective target (output) data into three distinct categories, allowing for a granular assessment of results in various data domains.
\subsection{Quantitative Analysis on BLIP}
    In the experimental setup involving the BLIP model, CLIPScore was selected as the quantitative evaluation metric \cite{hessel2022clipscorereferencefreeevaluationmetric}. The CLIPScore was computed for each iteration of the optimization process, across the three distinct categories of target data under consideration.
    \begin{table}[h]
    \centering
    \begin{tabular}{|c|c|c|c|}
    \hline
    Step & Simple Object & Multiple Entities & Abstract Concept \\
    \hline
    step 0 & 0.2079 & 0.2083 & 0.2471 \\
    \hline
    step 250 & 0.2118 & 0.2113 & 0.2496 \\
    \hline
    step 500 & 0.2155 & 0.2126 & 0.2493 \\
    \hline
    step 750 & 0.2161 & 0.2121 & 0.2541 \\
    \hline
    step 1000 & 0.2165 & 0.2116 & 0.2538 \\
    \hline
    \end{tabular}
    \caption{The CLIPScore is measured at steps 0, 250, 500, 750, and 1000 of the optimization process.}
    \label{BLIP-EVAL}
    \end{table}
\subsection{Quantitatve Analysis on Flux.1-dev}
    We applied the quantitative evaluation to the Flux.1-dev model. Here, the CLIP score served as our key metric, measuring the alignment between the optimized text generated by the inversion process and the target image \cite{hessel2022clipscorereferencefreeevaluationmetric}. We specifically examined three distinct categories of target images to assess the consistency of our results in various data domains.
    \begin{table}[h]
    \centering
    \begin{tabular}{|c|c|c|c|}
    \hline
    Step & Clear Object & Detailed Landscape & Artistic Image \\
    \hline
    step 0 & 0.1901 & 0.1817 & 0.1843 \\
    \hline
    step 25 & 0.1252 & 0.1897 & 0.1646 \\
    \hline
    step 50 & 0.1252 & 0.1885 & 0.1567 \\
    \hline
    step 75 & 0.1140 & 0.1813 & 0.2162 \\
    \hline
    step 100 & 0.0992 & 0.1813 & 0.2097 \\
    \hline
    \end{tabular}
    \caption{The CLIPScore is measured at steps 0, 25, 50, 75, and 100 of the optimization process.}
    \label{FLUX-EVAL}
    \end{table}
\subsection{Quantitative Analysis on Whisper-Large-V3}
    Similar to our previous analyses, we conducted a quantitative evaluation of the Whisper-Large-V3 model. For Whisper-Large-V3, the optimization process involves generating optimized audio from target text. Therefore, the Perceptual Evaluation of Speech Quality (PESQ) score was selected as our key quantitative metric, measuring the quality and similarity of the optimized audio against a reference \cite{941023}. We specifically examined three distinct categories of target text to assess the consistency of our results in various data domains.
    \begin{table}[h]
    \centering
    \begin{tabular}{|c|c|c|c|}
    \hline
    Step & Declarative Sentence & Complex Sentence & Emotive Sentence \\
    \hline
    step 0 & 1.06 & 1.02 & 1.03 \\
    \hline
    step 250 & 1.05 & 1.05 & 1.03 \\
    \hline
    step 500 & 1.05 & 1.11 & 1.03 \\
    \hline
    step 750 & 1.05 & 1.02 & 1.03 \\
    \hline
    step 1000 & 1.03 & 1.02 & 1.03 \\
    \hline
    \end{tabular}
    \caption{The PESQ score is measured at steps 0, 250, 500, 750, and 1000 of the optimization process.}
    \label{WHISPER-EVAL}
    \end{table}
\subsection{Quantitative Analysis on Chatterbox-TTS}
    Finally, we present the quantitative evaluation of the Chatterbox-TTS model. For the Chatterbox-TTS model, the optimization process generates optimized text from target audio. To assess the quality of this text, we selected the BERTScore, utilizing the Whisper-Large-V3 model as a reference transcription in each target audio for its robust transcription capabilities \cite{zhang2020bertscoreevaluatingtextgeneration}. The BERTScore was computed for each iteration of the optimization process across three distinct categories of target audio, allowing us to evaluate the consistency of our results in various data domains.
    \begin{table}[h]
    \centering
    \begin{tabular}{|c|c|c|c|}
    \hline
    Step & Clean Speech & Challenging Acoustics & Noisy Mixture \\
    \hline
    step 0 & 0.7607 & 0.7314 & 0.7401 \\
    \hline
    step 25 & 0.7607 & 0.7314 & 0.7401 \\
    \hline
    step 50 & 0.7607 & 0.7314 & 0.7401 \\
    \hline
    step 75 & 0.7607 & 0.7314 & 0.7401 \\
    \hline
    step 100 & 0.7607 & 0.7314 & 0.7401 \\
    \hline
    \end{tabular}
    \caption{The BERTScore is measured at steps 0, 25, 50, 75, and 100 of the optimization process.}
    \label{CHAT-EVAL}
    \end{table}

\section{Discussion}
    Our research investigates the invertibility of multimodal latent spaces, specifically through optimization-based methods. As our central hypothesis proposed that the multimodal latent spaces of task-specific models will not consistently support semantically meaningful and perceptually coherent inverse mapping through optimization-based methods, the experimental results align with our central hypothesis.
\subsection{Text-Image}
    In the Text-Image domain, our experiments with BLIP in generation task yielded promising initial results \cite{li2022blipbootstrappinglanguageimagepretraining}. When optimizing an image to match a target text ("A red apple on a wooden table"), we observed that the BLIP model, originally designed for image captioning, began to generate images that progressively aligned with the target caption. Both Adam and AdamW optimizers, irrespective of Gaussian noise or base image initialization, eventually produced images that BLIP itself accurately inferred as "a red apple on a wooden table" (Tables 1-4) \cite{kingma2017adammethodstochasticoptimization} \cite{loshchilov2019decoupledweightdecayregularization}. However, from a perceptual standpoint, the generated image was completely unsuccessful. The result demonstrates that BLIP's learned multimodal latent spaces are completely incapable of reconstructing visual semantics from textual goals, highlighting its implicit generative potential never works due to its nature as a discriminative model.
    \\\\
    The classification task with Flux.1-dev proved to be significantly more challenging \cite{blackforestlabs_2024_github}. Our objective was to infer the text embeddings that would produce a target image through a single-step inference. The optimization trajectory, visualized by the images generated in Figure 5, shows the degree of convergence towards the target image.
    \\\\
    However, the estimated tokens derived from the optimized embeddings (Tables 5 and 6) reveal a critical limitation. The cosine similarity scores for the closest vocabulary tokens were consistently low. (e.g., around 0.06-0.08 for token embeddings and 0.00-0.14 for the pooled embedding). These low scores indicate that while the optimization process might nudge the latent space towards generating the desired image, the resulting embeddings do not align strongly with any interpretable semantic tokens in the model's original vocabulary.
    \\\\
    The result of our investigation on Flux.1-dev suggests that while the image generation process in Flux.1-dev is robust. However, when applied for its inverse task, its internal textual latent space does not readily map back to clear, high-confidence token identities. Such consequences could be due to the highly compressed or abstract nature of latent space, or a significant discrepancy between the flexibility of forward mapping and the constraints of the inverse problem.
\subsection{Text-Audio}
    Our investigation of the Text-Audio domain revealed similar complexities. For Whisper-Large-V3 in a generation task, the optimization of a log-mel spectrogram to produce the target phrase "A red apple on a wooden table" showed progression (Figures 8-12) \cite{_2022_whisper} \cite{radford2022robustspeechrecognitionlargescale}. The transcriptions in Table 7 show that, through increasing optimization steps, Whisper eventually generated the exact target phrase. However, the reconstructed waveforms (Figure 13-17) visually confirm the persistent chaotic noise, which does not align with the textual goal. The reconstructed audio is a strong indicator that the model completely lacks the implicit generative potential required to synthesize coherent audio, despite its remarkable discriminative capabilities for transcription. Its internal latent spaces, while effective for recognition, do not translate into the robust capacity for audio generation.
    \\\\
    Attempting to try the classification task with Chatterbox-TTS presented considerable hurdles \cite{resembleai_2025_github}. The goal was to optimize text embeddings to generate a specific audio output ("A red apple on a wooden table"). While the mel spectrograms and waveforms (Figures 18-29) show the model's attempt to converge to the target audio, the estimated tokens (Table 8-11) reveal a lack of semantic interpretability, mirroring the issues faced with Flux.1-dev. The cosine similarity scores remained low, and the identified tokens often consisted of special characters, phonetic symbols (e.g., IPA characters such as \textbeltl, \textcrh), or obscure word fragments, rather than coherent semantic units. The results suggest that the latent space through which Chatterbox-TTS maps text to speech is highly specialized and not easily invertible to semantically meaningful text tokens.
\subsection{Overall Implications}
    Across both modalities, our findings suggest that optimization-based methods do not force models to produce output aligned with a target in a different modality. Task-specific classification models (e.g., image captioning, speech recognition) show no capacity for generative tasks, never successfully manipulating their input to achieve a perceptually meaningful output. Furthermore, when attempting to "classify" or infer semantics from task-specific generative models (e.g., inferring text from a text-to-image or text-to-speech model), the reconstructed embeddings consistently do not align with the model's own discrete vocabulary tokens in any semantically clear manner.
\section{Conclusion}
    This paper investigated the invertibility of multimodal latent spaces across different modalities (text, image, and audio) through the lens of optimization-based methods. Our central hypothesis assumed that the multimodal latent spaces of task-specific models will not consistently support semantically meaningful and perceptually coherent inverse mapping through optimization-based methods. Regardless of the varied results, our findings consistently proved the limitations of optimization-based methods, highlighting the critical need for further research into truly semantically rich and invertible multimodal latent spaces.
    \printbibliography
\end{document}